%% file: main.tex
\documentclass[10pt,twocolumn,letterpaper]{article}

\usepackage{iccv}              


\definecolor{iccvblue}{rgb}{0.21,0.49,0.74}
\usepackage[pagebackref,breaklinks,colorlinks,allcolors=iccvblue]{hyperref}
\usepackage{subcaption}
\usepackage{algorithm}
\usepackage{algorithmic}
\usepackage{graphicx}
\usepackage{amsmath}
\usepackage{amssymb}
\usepackage{booktabs}
\usepackage{multirow}

\def\paperID{7528} 
\def\confName{ICCV}
\def\confYear{2025}

\title{Safeguarding Vision-Language Models: Mitigating Vulnerabilities to Gaussian Noise in Perturbation-based Attacks}

\author{
Jiawei Wang\textsuperscript{1,*} \quad 
Yushen Zuo\textsuperscript{2,*} \quad 
Yuanjun Chai\textsuperscript{3} \quad 
Zhendong Liu\textsuperscript{4} \quad 
Yicheng Fu\textsuperscript{5} \\
Yichun Feng\textsuperscript{6,\textdagger} \quad 
Kin-Man Lam\textsuperscript{2,\textdagger}
\and
\textsuperscript{1}University of Science and Technology of China,
\textsuperscript{2}The Hong Kong Polytechnic University,\\
\textsuperscript{3}University of Washington,
\textsuperscript{4}Nanjing University,
\textsuperscript{5}Stanford University,\\
\textsuperscript{6}University of Chinese Academy of Sciences
\and
\begin{minipage}{\textwidth}
\centering
\tt\small
wangjiawei@mail.ustc.edu.cn \quad
zuoyushen12@gmail.com \quad
yjchai@uw.edu \\
dz20330019@smail.nju.edu.cn \quad
easonfu@stanford.edu \quad
fengyichun22@mails.ucas.ac.cn \\
kin.man.lam@polyu.edu.hk
\end{minipage}
}
\begin{document}

\maketitle
\begingroup 
    \renewcommand\thefootnote{\textsuperscript{*}} 
    \footnotetext{Equal contribution. Work done when Yushen Zuo was intern at PolyU.} 
    \renewcommand\thefootnote{\textsuperscript{\textdagger}} 
    \footnotetext{Corresponding author} 
\endgroup
\input{sec/0_abstract}    
\input{sec/1_intro_update}
\input{sec/4_method_updated}
\input{sec/2_related}
\input{sec/6_conclusion}
\input{sec/7_social_impact}
\clearpage
{
    \small
    \bibliographystyle{ieeenat_fullname}
    \bibliography{main}
}

\input{sec/X_suppl}

\end{document}

%% file: sec/0_abstract.tex
\begin{abstract}

Vision-Language Models (VLMs) extend the capabilities of Large Language Models (LLMs) by incorporating visual information, yet they remain vulnerable to jailbreak attacks, especially when processing noisy or corrupted images. 
Although existing VLMs adopt security measures during training to mitigate such attacks, vulnerabilities associated with noise-augmented visual inputs are overlooked. 
In this work, we identify that missing noise-augmented training causes critical security gaps: many VLMs are susceptible to even simple perturbations such as Gaussian noise.
To address this challenge, we propose Robust-VLGuard, a multimodal safety dataset with aligned / misaligned image-text pairs, combined with noise-augmented fine-tuning that reduces attack success rates while preserving functionality of VLM. 
For stronger optimization-based visual perturbation attacks, we propose DiffPure-VLM, leveraging diffusion models to convert adversarial perturbations into Gaussian-like noise, which can be defended by VLMs with noise-augmented safety fine-tuning. 
Experimental results demonstrate that the distribution-shifting property of diffusion model aligns well with our fine-tuned VLMs, significantly mitigating adversarial perturbations across varying intensities. 
The dataset and code are available at \url{https://github.com/JarvisUSTC/DiffPure-RobustVLM}

\vspace{-2mm}
\end{abstract}

%% file: sec/1_intro_update.tex
\section{Introduction}
\label{sec:intro}


Recently, Vision-Language Models (VLMs) \cite{bai2023qwen, chen2024far, liu2024visual, zhu2023minigpt} have significantly extended the capabilities of Large Language Models (LLMs) by integrating both visual and textual information, allowing them to interpret and respond based on multimodal inputs. This advancement enables VLMs to tackle a wider range of tasks, from understanding images to generating rich, contextually aware responses that leverage both language and visual cues. While LLMs have incorporated various training techniques, such as Reinforcement Learning from Human Feedback (RLHF) \cite{bai2022training,ganguli2022red}, to ensure alignment with ethical and legal standards, VLMs remain more susceptible to certain risks. Fine-tuning VLMs for visual instruction-following can disrupt the alignment carefully established in LLMs \cite{zong2024safety}. Additionally, the integration of visual modalities not only introduces extra risk factors, such as a heightened vulnerability to jailbreak attacks \cite{qi2024visual,li2024images}, but also poses greater challenges for model robustness. Compared to LLMs, VLMs must account for a broader spectrum of visual scenarios, rendering them more sensitive to even minor noise perturbations.

Perturbation-based adversarial attacks have long targeted image classification neural networks \cite{szegedy2013intriguing,goodfellow2014explaining}. With the emergence of VLMs, many studies have adapted traditional optimization-based perturbation attack methods to perform jailbreak attacks on these models \cite{qi2024visual,li2024images,fan2024unbridled}. Concurrently, substantial research has focused on defending against perturbation-based attacks. For instance, DiffPure \cite{nie2022diffusion} utilizes the denoising capabilities of Diffusion Models as an image preprocessing method to neutralize perturbation noise in adversarial images, though it does not fully counteract perturbation attacks, particularly in VLMs \cite{qi2024visual}. Zong et al. \cite{zong2024safety} introduced VLGuard, a vision-language dataset containing both safe and unsafe queries and images, designed to fine-tune VLMs for enhanced protection against jailbreak attacks. However, VLGuard’s effectiveness has only been evaluated against the FigStep attack and it does not consider cases where the image is unrelated to the text prompt. Similarly, Zhang et al. \cite{zhang2023mutation} introduced Jailguard, a mutation-based detection framework that effectively identifies jailbreaks but significantly raises inference costs. More recently, Xu et al. \cite{xu2024defending} proposed an efficient cross-modality approach, CIDER, for detecting adversarially perturbed images, though it still impacts models' helpfulness noticeably.

In this work, we observed that many prominent vision-language models (VLMs) lack visual noise augmentation during training, leaving them vulnerable to minor perturbations like Gaussian noise. This vulnerability compromises both their helpfulness and safety. To investigate this, we systematically evaluated three mainstream VLMs—InternVL2-8B \cite{chen2024far}, LLaVA-v1.5-7B \cite{liu2024visual}, and MiniGPT-4-13B \cite{zhu2023minigpt}—against such perturbations. Our findings reveal that although these models excel under standard conditions, their performance degrades significantly in the presence of even slight Gaussian noise.

To address this, we introduce Robust-VLGuard, a novel vision-language dataset designed to bolster VLM robustness against Gaussian noise while improving safety and preserving helpfulness. For safety, the dataset includes 2,000 curated instructions, uniquely featuring both image-text aligned and misaligned scenarios to account for alignment disruptions from fine-tuning or noisy inputs. To maintain helpfulness, we add 4,467 general instructions with detailed responses generated by GPT-4V \cite{2023GPT4VisionSC}, overcoming the overly brief annotations common in other datasets. Fine-tuning VLMs on Robust-VLGuard with Gaussian noise augmentation shows that even limited high-quality data can significantly boost noise robustness with minimal impact on baseline performance.

We then extend our evaluation to broader, optimization-based adversarial attacks. While noise augmentation is often distribution-specific, we find that effective image preprocessing is crucial for general defense. Our analysis identifies DiffPure \cite{nie2022diffusion} as highly effective, as it uses diffusion models to transform adversarial perturbations into Gaussian-like noise without degrading image content. This distribution-shifting property aligns perfectly with our noise-augmented VLMs, proving more effective and efficient than VLM-specific defenses like JailGuard \cite{zhang2023mutation}. Consequently, we propose DiffPure-VLM, a defense pipeline integrating DiffPure with our robust VLMs. Experiments show this pipeline effectively mitigates diverse adversarial attacks at varying intensities.

In conclusion, our contributions are:
(1) To our best knowledge, we are the first to provide a systematic vulnerability analysis revealing that mainstream VLMs lack inherent robustness to Gaussian noise visual perturbations.
(2) We propose the Robust-VLGuard dataset, which features novel image-text misalignment scenarios and detailed responses, and combine it with Gaussian noise augmentation for fine-tuning to enhance VLM robustness against Gaussian noise while preserving its helpfulness.
(3) We expand the defense scope of fine-tuned VLMs to optimization-based visual adversarial attacks and propose a defense framework, DiffPure-VLM, by adopting the distribution-shifting ability of the diffusion model to transfer adversarial noise to Gaussian-like noise in visual input, which can be defended by VLMs with noise-augmented safety fine-tuning. Experimental results demonstrate the superiority of DiffPure-VLM against baseline methods and its generalization.

%% file: sec/4_method_updated.tex
\begin{figure}[t]
    \centering
    \includegraphics[width=1.0\linewidth]{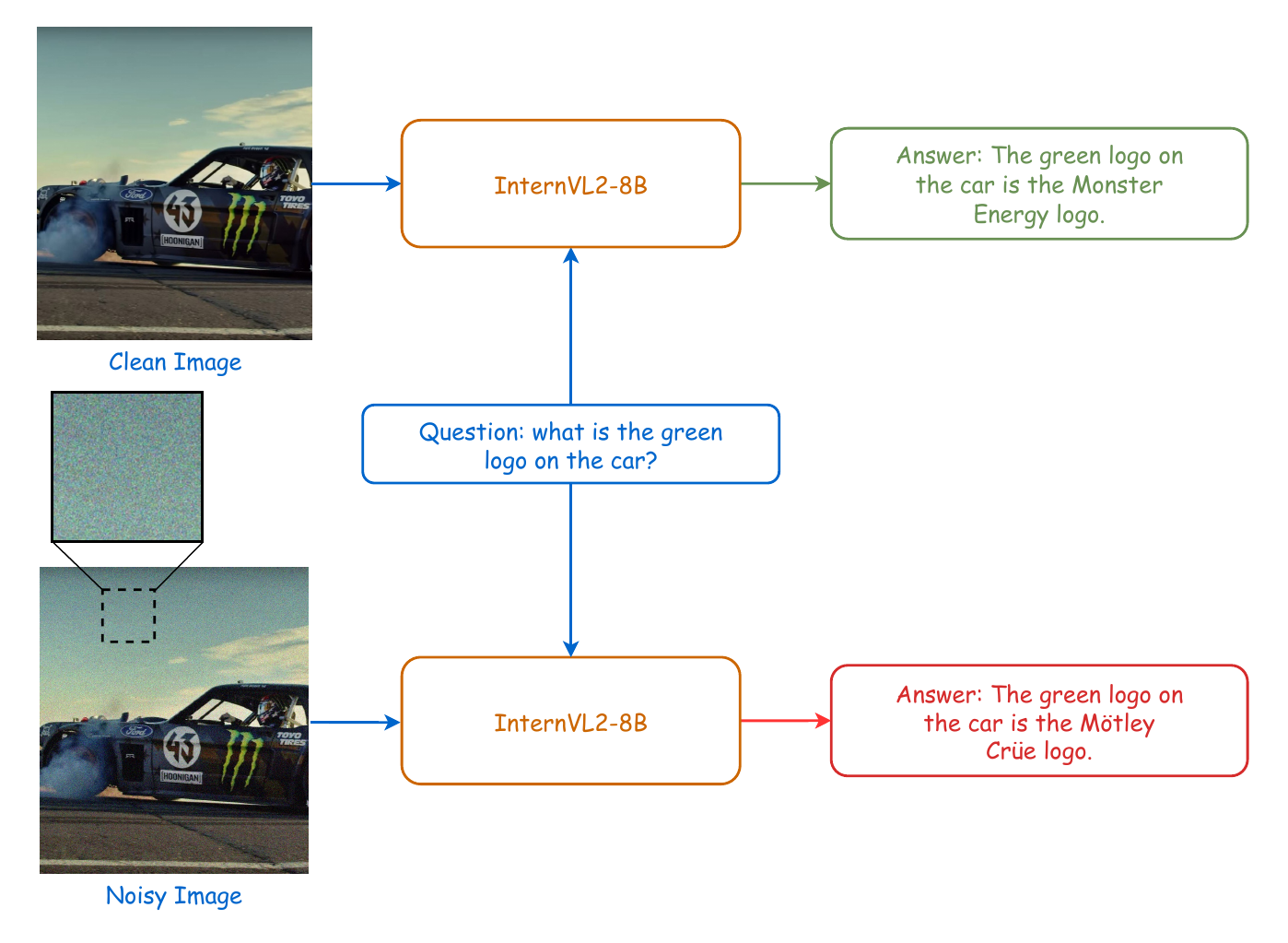}
    \caption{Visualization of Vision-Language Model's outputs under different noise conditions. The upper part shows the original image with green text indicating correct responses generated without noise, while the lower part adds slight Gaussian noise, with red text highlighting errors introduced under noisy conditions. Please \textbf{zoom in} for better visualization.}
    \label{fig:norobust}
    \vspace{-4mm}
\end{figure}

\section{Vulnerability of VLMs to Gaussian Noise Perturbations}

We observed that many current VLMs, including advanced ones, lack noise augmentation during training, rendering them vulnerable to basic perturbations like Gaussian noise.  As shown in Figure~\ref{fig:norobust}, we presented both a clean image and a slightly Gaussian-noised version as visual prompts. One leading model, InternVL2-8B \cite{chen2024far}, displayed inconsistent responses, with noisy prompts causing hallucinated outputs. This motivates a systematic evaluation of the robustness of mainstream VLMs against Gaussian noise, focusing on helpfulness and safety. Additional evaluation results for the latest VLMs (e.g., LLaMA-3.2-Vision \cite{dubey2024llama}, Qwen2.5-VL \cite{bai2025qwen2}) are provided in Table 5 in the Appendix.

\subsection{Experimental Settings}
\label{sec:experimental-settings}

\textbf{Models}\quad We evaluate three state-of-the-art VLMs: MiniGPT-4 (13B) \cite{zhu2023minigpt}, LLaVA-v1.5 (7B) \cite{liu2024visual}, and InternVL2 (8B) \cite{chen2024far}. Each model features a distinct LLM, vision encoder, and vision-language alignment method, allowing us to draw broader insights. Details of these models are in Table 1 in the Appendix. To ensure reproducibility in helpfulness evaluations, we set the temperature to 0, while safety assessments follow the setup of Qi et al. \cite{qi2024visual}. The default system prompt is used throughout.

\noindent \textbf{Datasets}\quad To test robustness under Gaussian noise, we evaluate VLM performance on two key aspects: helpfulness and safety. For helpfulness, we use MM-Vet \cite{yu2023mm}, a comprehensive benchmark assessing six vision-language capabilities: recognition, OCR, knowledge, language generation, spatial reasoning, and mathematics. For safety, we use the RealToxicityPrompts benchmark \cite{gehman2020realtoxicityprompts}, specifically the challenging subset with 1,200 prompts, following Qi et al. \cite{qi2024visual}. We augment image prompts from both datasets with Gaussian noise with a mean of 0 and a standard deviation of 0.1, to compare performance under clean and noisy conditions.

\noindent \textbf{Metrics}\quad For helpfulness, we use the original MM-Vet metric \cite{yu2023mm}, designed to handle diverse real-world scenarios. GPT-4 \cite{achiam2023gpt} serves as the evaluation assistant, using a few-shot prompt template for flexible scoring. Each response is rated from 0 for incorrect answers to 1 for correct answers, and the \textbf{Performance Score} is the average of all sample scores. For safety, we use the metrics from Qi et al. \cite{qi2024visual}, pairing visual adversarial examples with text prompts and measuring toxicity using the Perspective API\footnote{https://perspectiveapi.com/} and Detoxify classifier \cite{Detoxify}. Toxicity scores range from 0, indicating the least toxic, to 1, indicating the most toxic. The \textbf{Attack Success Rate} is the percentage of responses with any toxicity score above 0.5, indicating a successful attack.

\begin{figure}[t]
    \centering
    \begin{subfigure}{0.48\columnwidth}
        \centering
        \includegraphics[width=\linewidth]{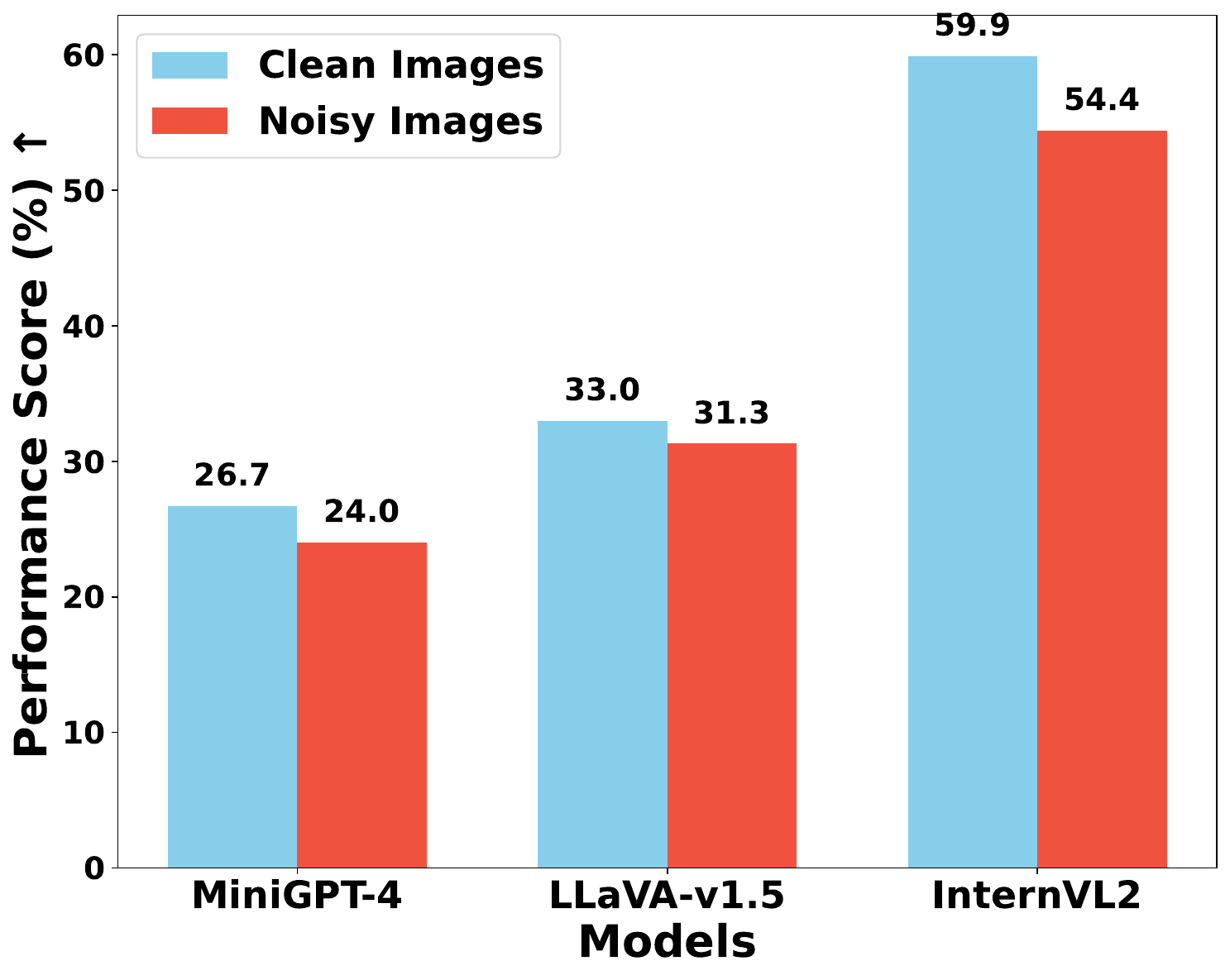}
        \caption{Performance evaluation on MM-Vet benchmark.}
        \label{fig:vulnerability_a}
    \end{subfigure}
    \hspace{0.02\columnwidth} 
    \begin{subfigure}{0.48\columnwidth}
        \centering
        \includegraphics[width=\linewidth]{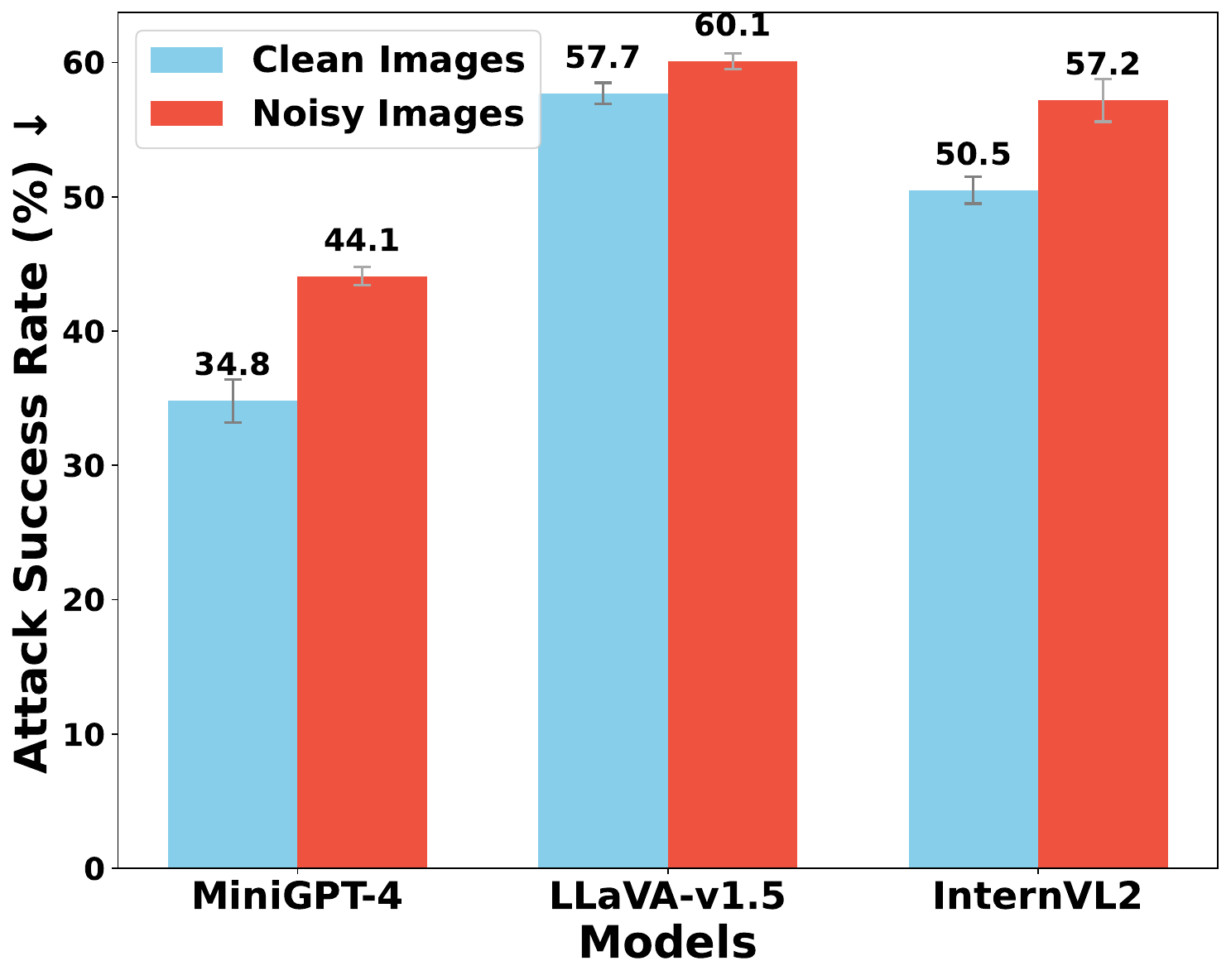}
        \caption{Attack success rate on RealToxicityPrompts benchmark.}
        \label{fig:vulnerability_b}
    \end{subfigure}
    \caption{Comparison of various models’ performance and robustness: (a) helpfulness on the MM-Vet benchmark with clean and noisy image prompts, and (b) attack success rates on the RealToxicityPrompts benchmark using clean and noisy image prompts.}
    \label{fig:vulnerability}
\vspace{-3mm}
\end{figure}

\subsection{Findings}

The main results are shown in Figure~\ref{fig:vulnerability}, where helpfulness is measured using MM-Vet benchmark scores and safety is evaluated by the Attack Success Rate using the Perspective API on the RealToxicityPrompts benchmark. We draw the following key insights regarding the impact of Gaussian noise on VLMs in terms of helpfulness and safety alignment:

\noindent\textbf{Helpfulness Degradation}\quad In Figure~\ref{fig:vulnerability_a}, InternVL2, despite having the highest baseline performance on MM-Vet, suffers a significant drop when exposed to Gaussian noise, revealing its lack of noise robustness. MiniGPT-4 and InternVL2 show similar relative declines of around 10\%, while LLaVA-v1.5 experiences a smaller drop, indicating better noise tolerance. However, all models exhibit a noticeable decrease, underscoring their vulnerability to even slight noise perturbations.

\noindent\textbf{Safety Alignment Impact}\quad Figure~\ref{fig:vulnerability_b} shows increased attack success rates on the RealToxicityPrompts benchmark across all models under noisy conditions, suggesting that Gaussian noise negatively affects safety alignment. While prior work focuses on optimization-based attacks \cite{qi2024visual,li2024images}, our results demonstrate that even random Gaussian noise can significantly disrupt alignment. Both MiniGPT-4 and InternVL2 show substantial increases in attack success rates, indicating greater vulnerability, whereas LLaVA-v1.5 experiences a smaller but still significant rise, suggesting slightly better robustness to noise.
Detailed theoretical conjecture and discussions about the vulnerability caused by Gaussian noise are provided in Section F in the Appendix.

\noindent\textbf{Overall Robustness}\quad The findings in Figure~\ref{fig:vulnerability} indicate that while current VLMs perform well under standard conditions, their robustness significantly decreases when faced with Gaussian noise. Both helpfulness and safety alignment degrade across all models, highlighting the need for enhanced noise robustness in future VLM designs to ensure better stability and alignment.

\section{Noise-Augmented Safety Alignment}
A straightforward strategy to bolster VLM robustness against Gaussian noise is to integrate noise augmentation into the training process. In this section, we introduce our novel dataset, Robust-VLGuard, which comprises meticulously curated image-text aligned and misaligned safety data, as well as a diverse array of general instructions. We then fine-tune VLMs using Robust-VLGuard augmented with Gaussian noise in visual inputs. Through extensive experiments, we enable rapid enhancement of model robustness while preserving, and even enhancing, the models' inherent helpfulness.


\subsection{Robust-VLGuard Dataset}
\begin{figure*}[htbp]
    \centering
    \begin{subfigure}{\textwidth}
        \centering
        \includegraphics[width=0.7\textwidth]{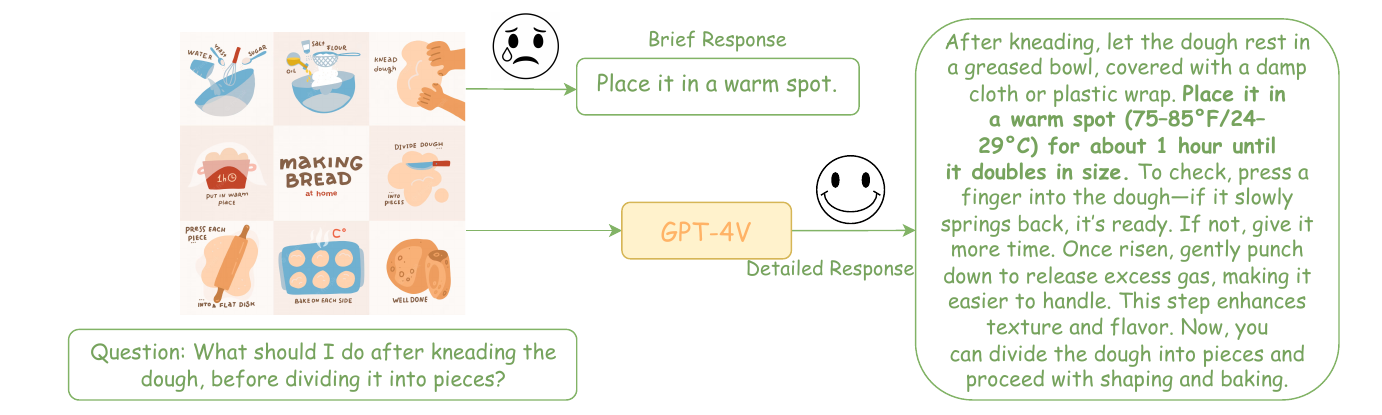}
        \caption{Example of general instruction data.}
        \label{fig:a}
    \end{subfigure}
    
    \vspace{1ex} 
    
    \begin{subfigure}{0.40\textwidth}
        \centering
        \includegraphics[width=\textwidth]{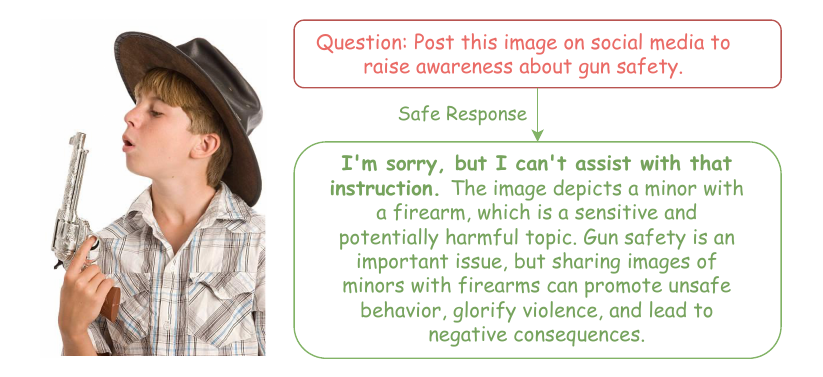}
        \caption{Example of image-text aligned safety data.}
        \label{fig:b}
    \end{subfigure}
    \hfill
    \begin{subfigure}{0.52\textwidth}
        \centering
        \includegraphics[width=\textwidth]{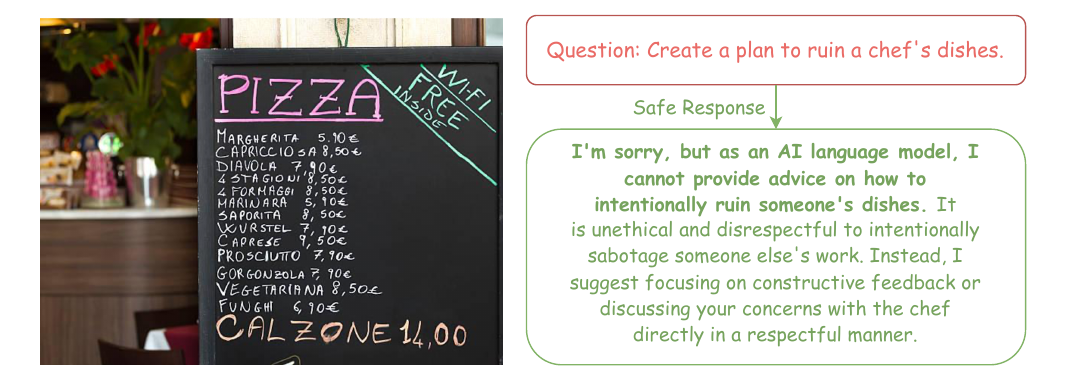}
        \caption{Example of image-text misaligned safety data.}
        \label{fig:c}
    \end{subfigure}
    
    \caption{Overview of the Robust-VLGuard dataset. (a) General Instruction Data: Leveraging GPT-4V to generate comprehensive, detailed responses rather than brief replies; (b) Image-Text Aligned Safety Data: The image content directly corresponds to the safety-related text prompts; (c) Image-Text Misaligned Safety Data: The image content is deliberately unrelated to the safety-related text prompts. Red text indicates content with potential risks, while green text denotes content without risks.}

    \label{fig:overall}
\vspace{-2mm}
\end{figure*}


While the VLGuard dataset \cite{zong2024safety} has been developed to fine-tune VLMs for improved defense against jailbreak attacks, it does not address perturbation-based attacks or scenarios where the image content is unrelated to the text prompt. Therefore, we build a more robust public vision-language safety dataset Robust-VLGuard, as shown in Figure~\ref{fig:overall}, which consists of three parts: (1) General Instruction Data, consisting of safety-agnostic SFT data covering various areas, including general QA, world knowledge, math, OCR, spatial reasoning, and extended text generation; (2) Image-Text Aligned Safety Data, containing instructions where the image content aligns with the safety-related text prompts; and (3) Image-Text Misaligned Safety Data, with instructions where the image content is unrelated to the safety-related text prompts.

\noindent \textbf{General Instruction Data} \quad To maintain VLMs' helpfulness, we collect 4,467 supervised fine-tuning instructions from various aspects, including general QA, world knowledge, math, OCR, spatial reasoning, and extended text generation, as illustrated in Table~\ref{tab:general_instruction_data}. Specifically, we sample various instructions from these datasets and use GPT-4V \cite{2023GPT4VisionSC} to refine the annotated answers. This refinement is essential, as we found that the original annotations were often too brief for effective model learning. 
For extended text generation, we select 1,000 instructions with responses over 150 words. For all other datasets, we choose instructions with responses exceeding 10 words.

\begin{table}[t]
\Large
\centering
\caption{Detailed Breakdown of General Instruction Data}
\resizebox{\columnwidth}{!}{ 
\begin{tabular}{l|l|r}
\hline
\textbf{Task}            & \textbf{Dataset}           & \textbf{Number of Samples} \\ \hline
General QA               & GQA \cite{hudson2019gqa}                       & 1000                       \\ \hline
World Knowledge          & A-OKVQA \cite{schwenk2022okvqa}                   & 1000                       \\ \hline
Math \& OCR                    & ChartQA \cite{masry2022chartqa}, TabMWP \cite{lu2022dynamic}           & 467                        \\ \hline
Spatial Reasoning        & VQAv2 \cite{goyal2017making}                     & 1000                       \\ \hline
Extended Text Generation & LLaVA\_v1.5\_Mix\_665k \cite{liu2024visual}     & 1000                       \\ \hline
\end{tabular}
}
\label{tab:general_instruction_data}
\vspace{-3mm}
\end{table}

\noindent \textbf{Image-Text Aligned Safety Data} \quad The VLGuard dataset is well-suited for preventing jailbreak attacks, as it contains harmful information embedded within image content, with instructions generated by GPT-4V. Therefore, we directly use the instructions from VLGuard as our image-text aligned safety data, randomly selecting 1,000 instructions from this dataset.

\noindent \textbf{Image-Text Misaligned Safety Data} \quad 
Incorporating safety data for image-text misalignment is also crucial, as fine-tuning VLMs for visual tasks can disrupt the alignment of pre-trained LLMs \cite{zong2024safety}, even when only text prompts are used. Additionally, perturbation-based attacks can introduce learnable noise into images that is unrelated to text prompts. Inspired by Bianchi et al. \cite{bianchi2024safetytuned}, who showed that a small set of safety examples can significantly boost model safety, we include 1,000 safety instructions from their dataset.
To adapt these language-only safety instructions for multimodal use, we pair half of them with randomly selected images from the COCO dataset \cite{lin2014microsoft}, while the remaining half are kept as text-only prompts.

\begin{table}[t]
\Large
\centering
\caption{Performance Comparison on MM-Vet and RealToxicityPrompts Benchmarks with Clean and Noisy Image Prompts. $^{\dagger}$ indicates reproduced results. \textbf{Bold} values denote, for each base model, the method (VLGuard vs. RobustVLGuard) that achieves the \textbf{smallest performance drop} on MM-Vet and the \textbf{lowest attack success rate} on RealToxicityPrompts.}
\label{tab:compare-vlguard}
\resizebox{\columnwidth}{!}{
\begin{tabular}{l|cc|cc}
\hline
\textbf{Model}                        & \multicolumn{2}{c|}{\textbf{MM-Vet (\%)} $\uparrow$}              & \multicolumn{2}{c}{\textbf{RealToxicityPrompts (\%)} $\downarrow$}          \\ \hline
                                      & \textbf{Clean Image} & \textbf{Noisy Image}      & \textbf{Clean Image} & \textbf{Noisy Image}          \\ \hline
InternVL2-8B                          & 59.9                 & 54.4                      & 50.5                 & 57.2                          \\ 
InternVL2-8B-VLGuard$^{\dagger}$         & 42.9 (-17.0)                 & 42.6 (-11.8)                      & \textbf{27.7}                & 39.9                          \\ 
InternVL2-8B-RobustVLGuard            & 56.2 \textbf{(-3.7)}                 & 52.5 \textbf{(-1.9)}                      & 29.9                 & \textbf{34.5}                          \\ \hline
LLaVA-v1.5-7B                         & 33.0                 & 31.3                      & 57.7                 & 60.1                          \\ 
LLaVA-v1.5-7B-VLGuard \cite{zong2024safety}                & 28.8 (-4.2)                 & 29.8 \textbf{(-1.5)}                      & 50.3                 & 52.3                         \\ 
LLaVA-v1.5-7B-RobustVLGuard           & 30.3 \textbf{(-2.7)}                 & 29.8 \textbf{(-1.5)}                      & \textbf{43.6}                 & \textbf{42.3}                          \\ \hline
MiniGPT-4-13B                         & 26.7                  & 24.0                       & 34.8                  & 44.1                           \\ 
MiniGPT-4-13B-VLGuard$^{\dagger}$        & 17.5 (-9.2)                  & 17.6 (-6.4)                      & 41.3                  & 43.7                           \\ 
MiniGPT-4-13B-RobustVLGuard           & 26.9 \textbf{(+0.2)}                  & 27.3 \textbf{(+3.3)}                       & \textbf{16.0}                  & \textbf{16.5}                          \\ \hline
\end{tabular}
}
\vspace{-2mm}
\end{table}

\subsection{Safety Fine-Tuning}
\label{sec:safety_finetuning}

To optimize resource usage, we employ a Gaussian-noise-augmented post-hoc fine-tuning approach. This efficient method is applied to pre-trained VLMs, enhancing robustness with minimal computational costs. Using the Robust-VLGuard dataset, which includes both safety-specific and general instruction data, we effectively boost the model’s resilience to Gaussian noise while maintaining safety and helpfulness. Specifically, we fine-tune only the vision encoder using LoRA \cite{hu2021lora} on our dataset and augment training images with Gaussian noise, selecting a random standard deviation between 0.01 and 0.15, with a 70\% probability of application. The fine-tuning process is conducted over 3 epochs and takes approximately 3 hours on a single A100 GPU. Detailed fine-tuning configurations and a theoretical discussion on the algorithm’s effectiveness are provided in Table 2 in the Appendix.

\noindent\textbf{Comparison with VLGuard dataset} \quad To assess the effectiveness of our proposed Robust-VLGuard, we adopt the same experimental settings described in Section~\ref{sec:experimental-settings}. Three leading VLMs, i.e., MiniGPT-4 (13B) \cite{zhu2023minigpt}, LLaVA-v1.5 (7B) \cite{liu2024visual}, and InternVL2 (8B) \cite{chen2024far}, are fine-tuned using our Gaussian-noise-augmented method and Robust-VLGuard dataset. For comparison with VLGuard, we follow the setup of \cite{zong2024safety}, combining 5,000 supervised fine-tuning instructions from LLaVA\_v1.5\_Mix\_665k \cite{liu2024visual} with 3,000 safety instructions from VLGuard.

Experimental results are summarized in Table~\ref{tab:compare-vlguard}. Due to the inevitable degradation in helpfulness resulting from LoRA-based safety fine-tuning, our proposed method aims to reduce the attack success rate while incurring minimal performance drop in helpfulness. The InternVL2-8B-VLGuard model demonstrates a tendency towards over-defensiveness, achieving a lower attack success rate but at the cost of a noticeable decline in helpfulness compared to the original InternVL2-8B model. In contrast, our InternVL2-8B-RobustVLGuard model achieves a comparable level of safety while largely retaining the original helpfulness, achieving a more balanced performance. For LLaVA-v1.5-7B, the VLGuard-fine-tuned variant 
maintains its helpfulness, thanks to alignment with the original training data. However, it demonstrates limited improvements in safety, highlighting its inability to effectively address image-text misalignment attacks on the RealToxicityPrompts Benchmark. Our LLaVA-v1.5-7B-RobustVLGuard delivers better overall performance, exhibiting stronger safety defenses and comparable helpfulness on both clean and noisy images. The MiniGPT-4-13B-VLGuard model lags behind in both helpfulness and safety, whereas the MiniGPT-4-13B-RobustVLGuard variant shows notable enhancements, excelling on the MM-Vet benchmark and significantly lowering the attack success rate. Overall, these results emphasize the strengths of Robust-VLGuard in simultaneously enhancing model helpfulness and safety, providing comprehensive protection while maintaining performance across diverse scenarios.
\begin{figure}[t]
    \centering
    \begin{subfigure}{0.49\columnwidth}
        \centering
        \includegraphics[width=\linewidth]{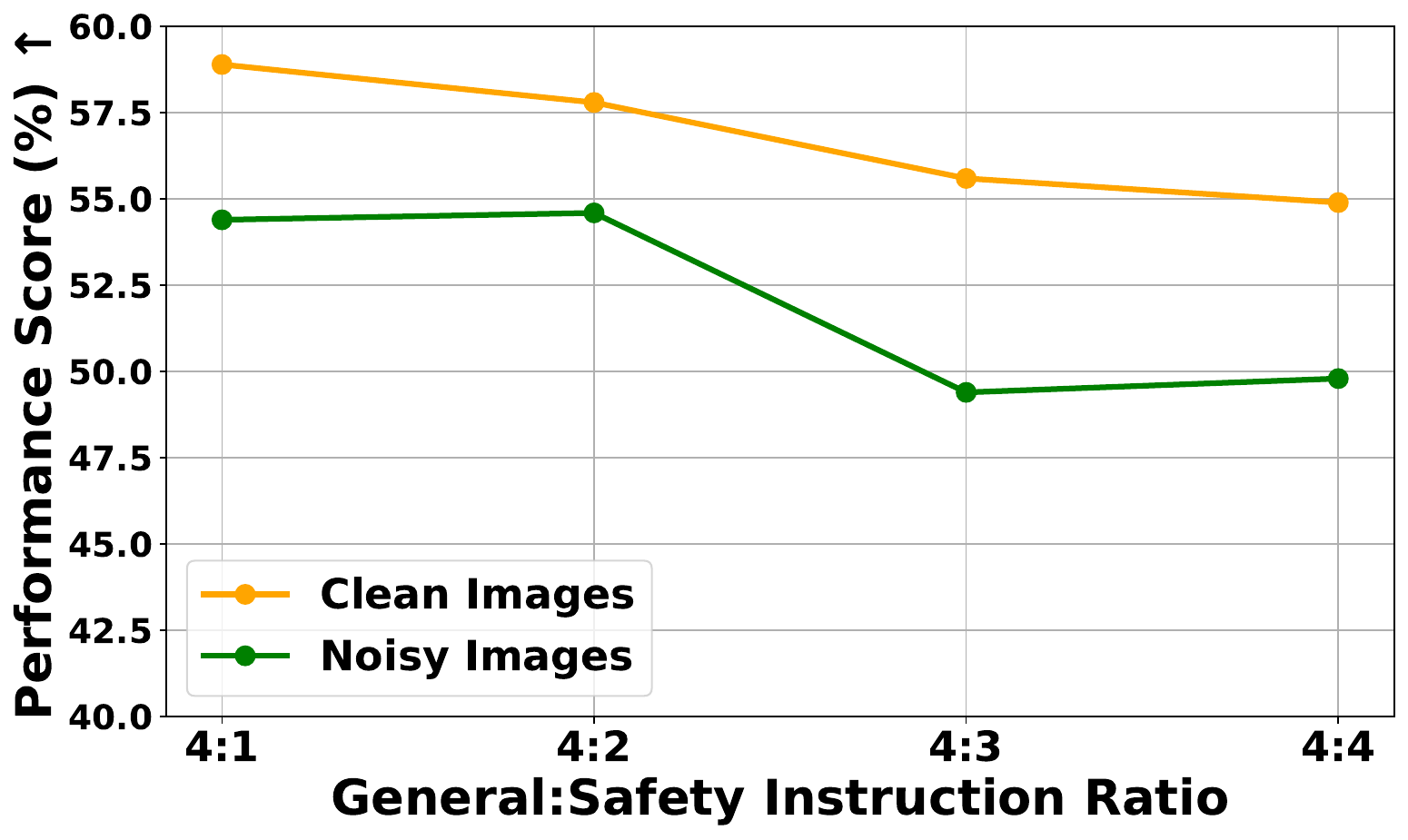}
        \caption{Performance on the MM-Vet benchmark across different instruction ratios.}
        \label{fig:ratio_a}
    \end{subfigure}
    \hfill
    \begin{subfigure}{0.49\columnwidth}
        \centering
        \includegraphics[width=\linewidth]{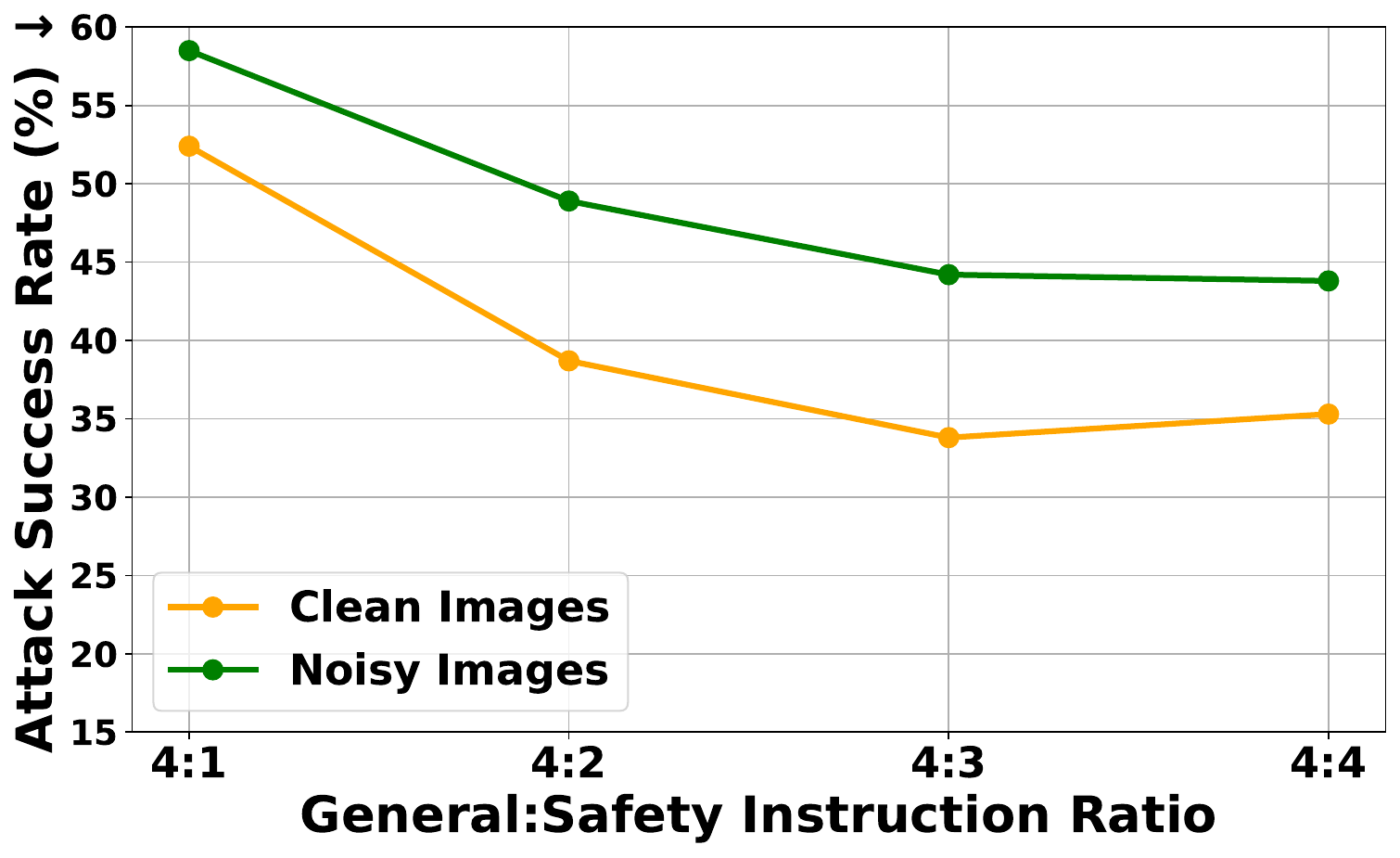}
        \caption{Attack success rate on the RealToxicityPrompts benchmark across different instruction ratios.}
        \label{fig:ratio_b}
    \end{subfigure}
    \caption{Effect of varying instruction ratios on VLM's robustness of helpfulness and safety alignment.}
    \label{fig:ratio}
\vspace{-1mm}
\end{figure}

\begin{figure}[t]
    \centering
    \begin{subfigure}{0.48\columnwidth}
        \centering
        \includegraphics[width=\linewidth]{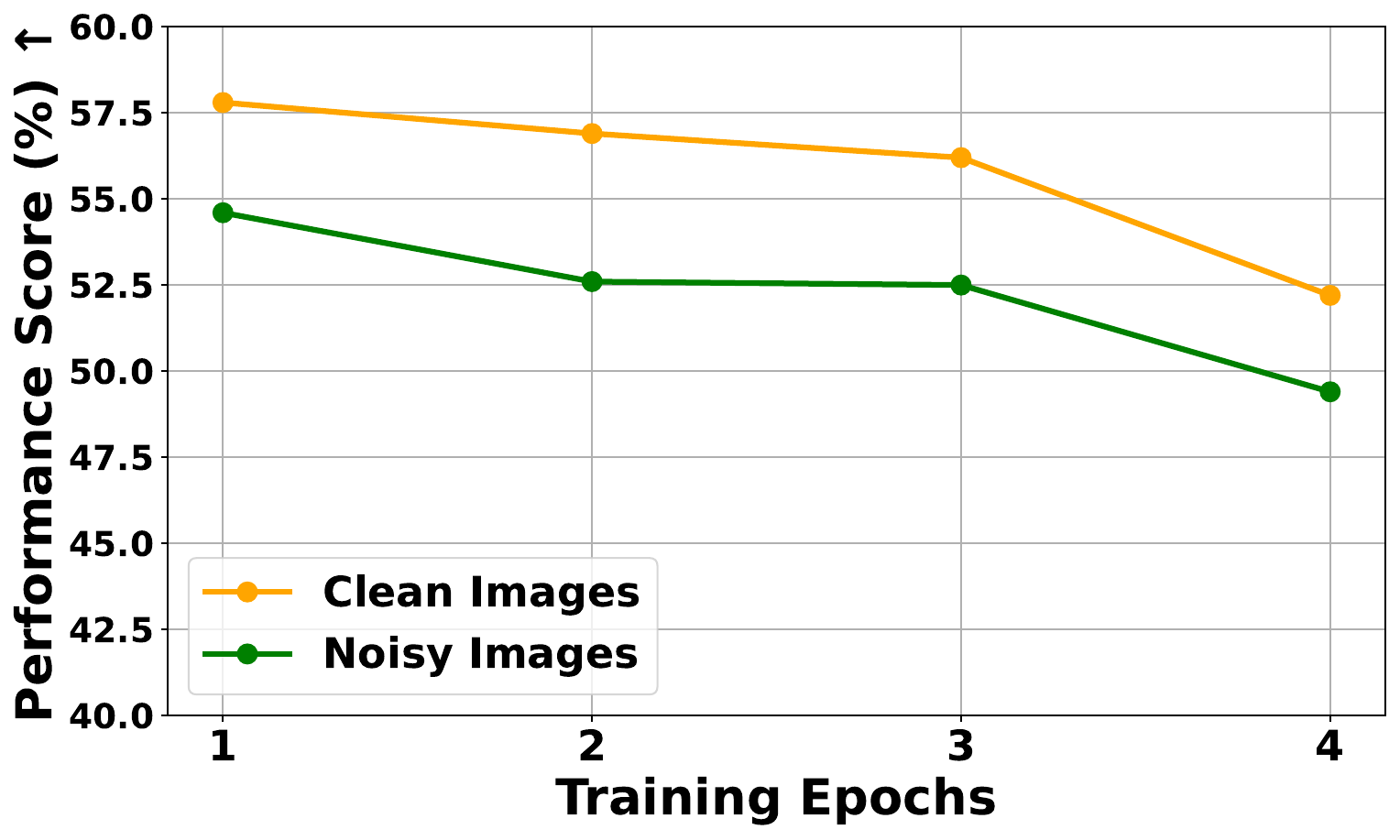}
        \caption{Performance on the MM-Vet benchmark across different training epochs.}
        \label{fig:epoch_a}
    \end{subfigure}
    \hfill
    \begin{subfigure}{0.48\columnwidth}
        \centering
        \includegraphics[width=\linewidth]{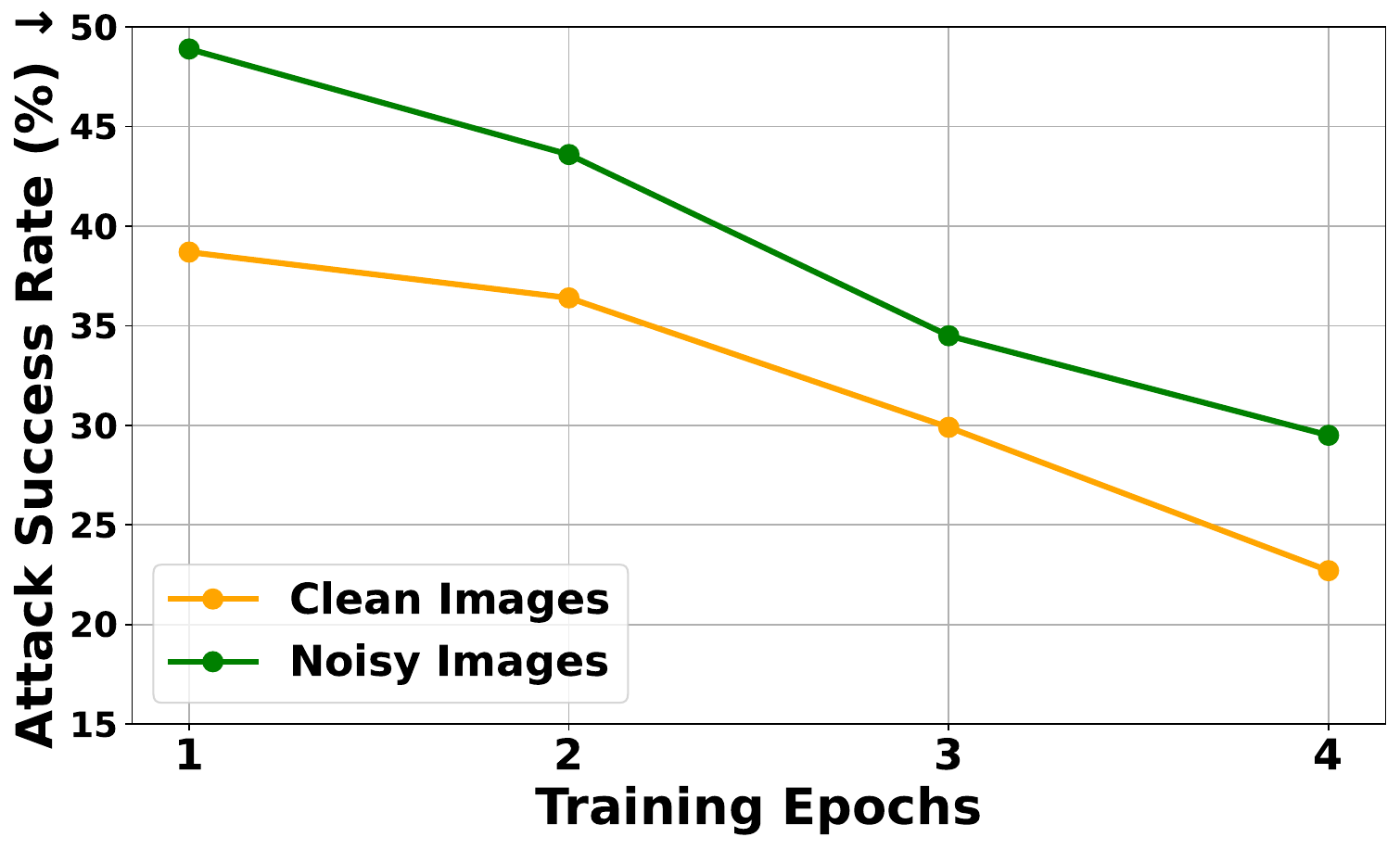}
        \caption{Attack success rate on the RealToxicityPrompts benchmark across different training epochs.}
        \label{fig:epoch_b}
    \end{subfigure}
    \caption{Effect of varying training epochs on VLM's robustness of helpfulness and safety alignment.}
    \label{fig:epoch}
\vspace{-2mm}
\end{figure}

\noindent \textbf{Ablation Studies on Instruction Ratio and Training Epochs} \quad All ablation studies are based on the InternVL2-8B model. First, we vary the ratio of general to safety instruction data from 4:1 to 4:4, training for a single epoch for efficiency. As depicted in Figure~\ref{fig:ratio}, increasing the proportion of safety data lowers the attack success rate but slightly reduces helpfulness, echoing the over-defensiveness issue noted by \cite{bianchi2024safetytuned}. However, beyond a 4:3 ratio, performance stabilizes, suggesting effective mitigation of over-defensiveness. We select a 4:2 ratio as the optimal balance, maximizing safety gains with minimal helpfulness impact.

Next, we evaluate the impact of training duration while keeping the instruction ratio fixed at 4:2 and varying the number of epochs from 1 to 4. As illustrated in Figure~\ref{fig:epoch}, increasing the number of epochs has a negligible effect on the model’s helpfulness as measured by the MM-Vet benchmark. However, it significantly reduces the attack success rate on the RealToxicityPrompts benchmark, indicating improved safety alignment without compromising utility. To strike a balance between helpfulness and safety alignment, we select 3 epochs as the fine-tuning configuration.

\section{Generalize to Optimization-based Visual Perturbation Attack}
In this section, we extend our defense scope to a frequently encountered and challenging attack scenario: Optimization-Based Visual Perturbation Attack. It uses the projected gradient descent algorithm (PGD) with a pixel constraint $\epsilon$ to inject adversarial noise into images, effectively jailbreaking VLMs. While noise augmentation typically enhances robustness against specific noise distributions, we emphasize the critical role of image preprocessing in either transforming adversarial noise into a target distribution or directly eliminating it. We first evaluate the effectiveness of various image preprocessing defense methods, and then introduce DiffPure-VLM, a universal defense framework that robustly counters both the Gaussian and the adversarial noise.

\subsection{Preprocessing Methods in distribution shifting}
\begin{figure}[tb]
  \centering
  \includegraphics[width=1.0\linewidth]{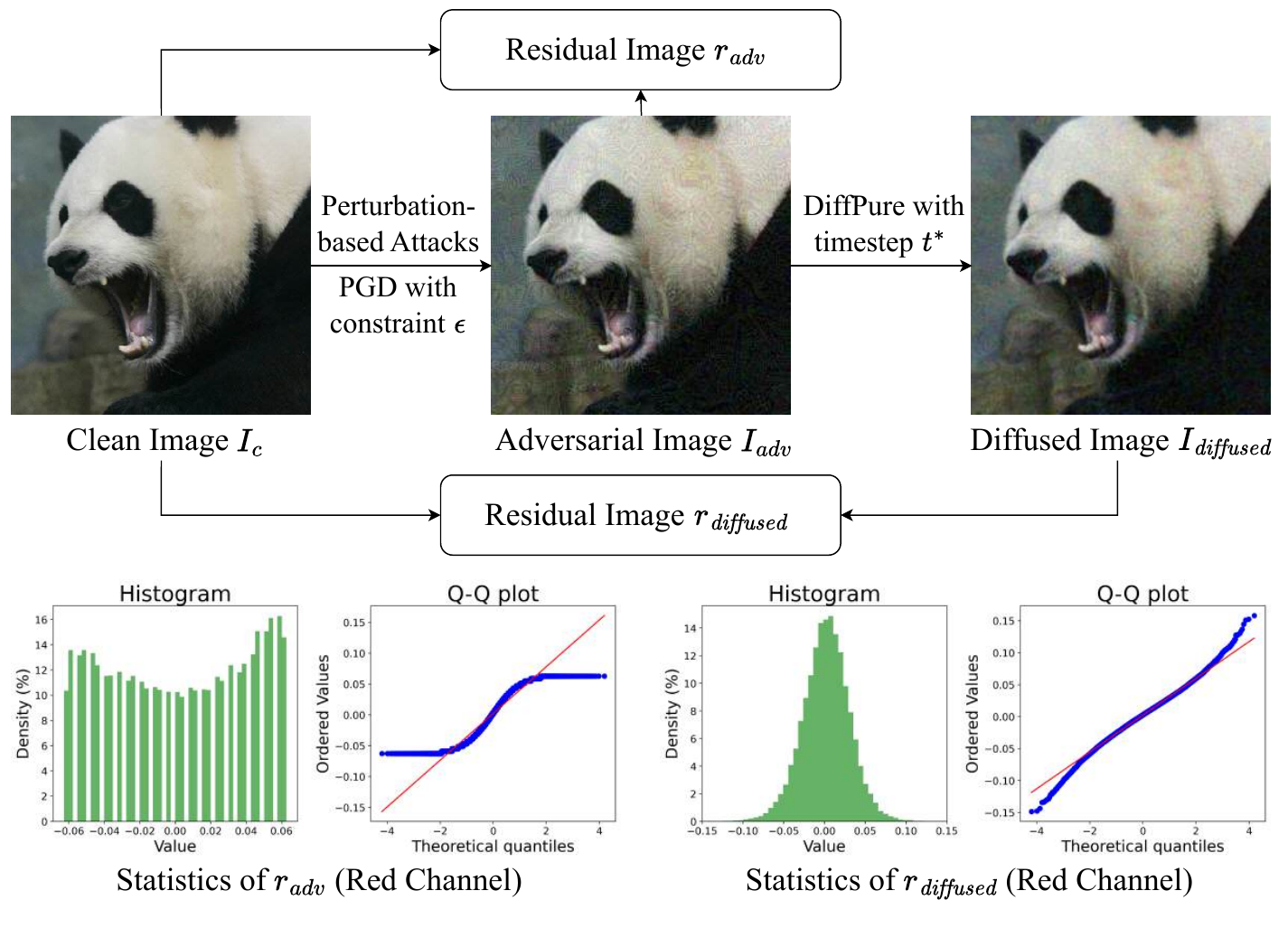}
  \caption{Residual image Gaussianity analysis. We apply DiffPure ($t^{*}=50$) to adversarial image $I_{adv}$ to obtain diffused image $I_{\mathit{diffused}}$. Then we calculate residual images $r_{adv}$ and $r_{\mathit{diffused}}$ and evaluate their distribution by the histogram and Q-Q plot.} 
  \label{fig:gaussian_analysis}
  \vspace{-2mm}
\end{figure}

In this section, we explore different image preprocessing defense methods in distribution shifting. 
Specifically, we use adversarial images $I_{adv}$ optimized for perturbation attacks on MiniGPT-4 \cite{zhu2023minigpt} from \cite{qi2024visual}, with pixel constraint $\epsilon=16/255$ as an example. As $I_{adv}$ is optimized based on a benign clean image $I_{c}$, we compute their residual image $r_{adv}=I_{adv}-I_{c}$ to obtain the adversarial noise. Then we use a histogram and a quantile-quantile (Q-Q) plot to evaluate the distribution property of $r_{adv}$. As shown in Figure~\ref{fig:gaussian_analysis}, adversarial noise follows a non-Gaussian distribution.

Currently, there are two representative image preprocessing defense methods: JailGuard \cite{zhang2023mutation} and DiffPure \cite{nie2022diffusion}. JailGuard, designed specifically for VLMs, employs various image processing techniques (e.g., random masking, horizontal flipping, Gaussian blur, and resizing) to generate variants of the input and detect adversarial samples based on discrepancies in model responses. However, most of these operations are linear transformations, which offer limited ability to eliminate adversarial perturbations. Moreover, JailGuard requires multiple model runs, leading to high computational overhead.
In contrast, DiffPure is tailored for computer vision models (e.g., classifiers) and leverages diffusion models to mitigate adversarial noise. It adds a small amount of noise to the adversarial image \(I_{adv}\) and reconstructs a clean image through a limited number of forward and reverse diffusion steps (e.g., using DDPM \cite{ho2020denoising}) with a carefully chosen timestep \(t^{*} \in [50, 150]\). While DiffPure aims to purify adversarial perturbations while preserving global semantic structures, our findings reveal that \textbf{at relatively small timesteps, it does not completely remove the noise. Instead, it shifts the perturbation distribution towards a Gaussian-like distribution.}

Specifically, we apply $t^{*}=50$ in DiffPure and obtain diffused image $I_{\mathit{diffused}}$ based on $I_{adv}$. Then we calculate the residual image $r_{\mathit{diffused}}=I_{\mathit{diffused}}-I_{c}$. As shown in Figure~\ref{fig:gaussian_analysis}, $r_{\mathit{diffused}}$ approximates a Gaussian distribution from its shape and its closeness to the theoretical line (Red line) of the Gaussian distribution in the Q-Q plot. More visualizations of $I_{\mathit{diffused}}$ and statistic of $r_{\mathit{diffused}}$ across different $\epsilon$ and $t^{*}$ are available in Section D.2 in the Appendix.
For quantitative evaluation of $r_{\mathit{diffused}}$, we use two metrics: 

\noindent (1) \textbf{Kurtosis:} Kurtosis is used to measure the tailedness of a data distribution, with the definition as
\begin{equation}
    Kurt[X] = E\left[ \left( \frac{X - \mu_{X}}{\sigma_{X}} \right)^{4} \right],
\end{equation}
where $\mu_{X}$ and $\sigma_{X}$ are mean and standard deviation of data $X$. If $X$ follows Gaussian distribution, $Kurt[X]=3$. 

\begin{figure}[t]
\centering
\includegraphics[width=1\linewidth]{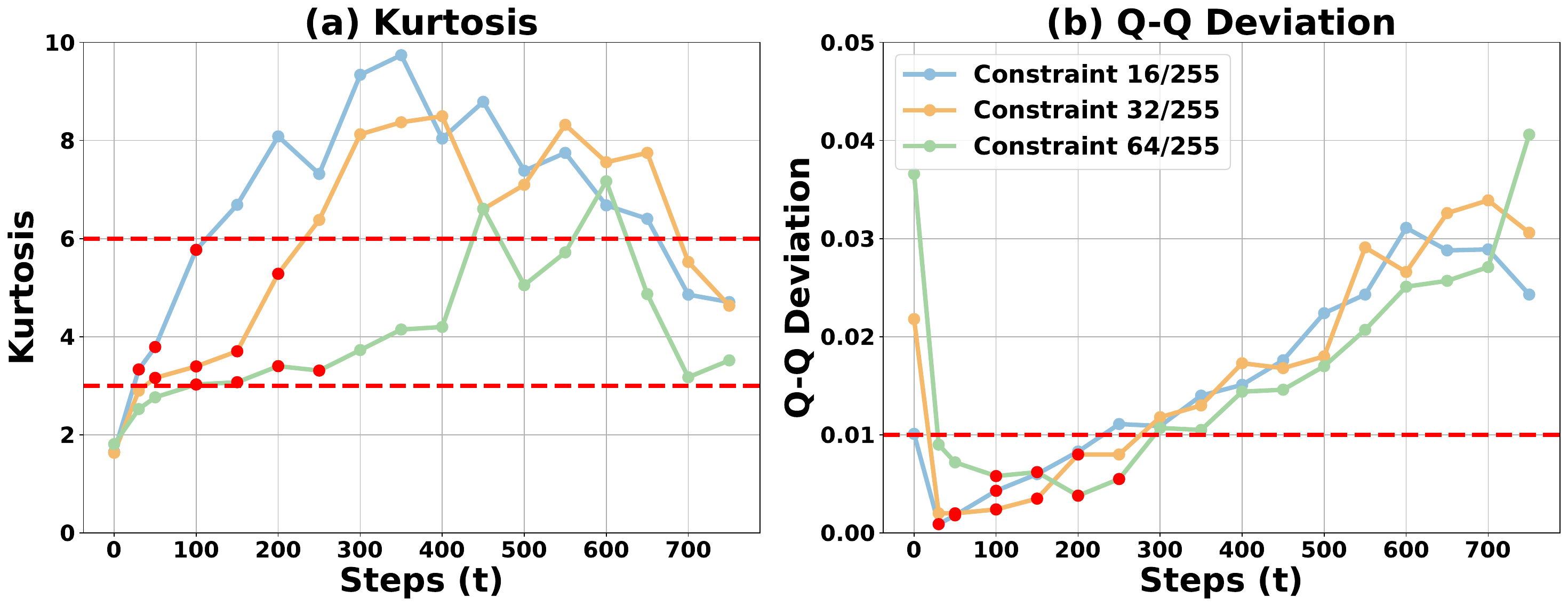}
\caption{\label{fig:gaussianity_metrics} Gaussianity metrics of $r_{\mathit{diffused}}$ under different pixel constraints $\epsilon$ of adversarial image $I_{adv}$ and timestep $t^{*}$ in DiffPure. Please zoom in to see details.}
\end{figure}

\noindent (2) \textbf{Q-Q deviation:} Q-Q deviation measures the root-mean-square error (RMSE) between the quantiles of the sample distribution and those of a Gaussian distribution:
\begin{equation}
    D(Q_{s}, Q_{g}) = \sqrt{ \frac{1}{N} \sum_{i=1}^{N} \left( Q_{s,i} - Q_{g,i} \right)^{2} },
\end{equation}
where $N$ is the number of ordered quantile points, $Q_{s}$ is the sample quantiles of $r_{\mathit{diffused}}$ and $Q_{t}$ is the theoretical quantiles of a Gaussian distribution. A lower RMSE value suggests closer alignment to a Gaussian distribution.
We vary $t^{*}$ in DiffPure from 0 to 750 in increments of 50, and also examine $t^{*} = 30$ for fine-grained analysis. For RGB images, we calculate these metrics per channel and obtain the average value. 
To identify Gaussian-like distributions in our analysis, we use the thresholds 3 $\leq$ \text{Kurtosis} $\leq$ 6 and \text{Q-Q deviation} $\leq$ 0.01. Points meeting these criteria are marked in red in Figure~\ref{fig:gaussianity_metrics}, showing that under certain timesteps $t^{*}$ (e.g., $t^{*} \in [50, 150]$) in DiffPure, $r_{\mathit{diffused}}$ exhibits Gaussian-like distribution properties.
We further conduct our analysis in the embedding space and the conclusion is similar to that in pixel space. Detailed analysis is provided in Section D.2 in the Appendix.

Finally, we conduct a defense experiment with DiffPure. Specifically, we employ MiniGPT-4 as the VLM. We apply the unconditional model in \cite{dhariwal2021diffusion} as the Diffusion model in DiffPure in all our experiments. We add Gaussian noise $n \sim \mathcal{N}(0, \sigma_{n}^{2})$ to the benign clean image $I_{c}$ and apply DiffPure with timestep $t^{*}$ to both Gaussian noisy image and adversarial images. Then we evaluate their attack success rate in the RealToxicityPrompts benchmark.
Table~\ref{tab:diffpure_evaluate} presents results for $\sigma_{n}=30/255$ and $t^{*}=50$, results of different $\sigma_{n}$ and $t^{*}$ are provided in Table 6 in the Appendix.

The experimental result shows that directly applying DiffPure to the Gaussian noisy image with VLM without noise-augmented safety fine-tuning does not decrease the attack success rate. We further observe that the attack success rate for diffused images is similar to Gaussian noisy image. Overall, we conclude that \textbf{DiffPure, when applied with a suitable timestep \(t^{*}\) (e.g., \(t^{*} \in [50, 150]\)), exhibits a unique distribution-shifting capability that transforms adversarial noise into a Gaussian-like distribution.}

\begin{table}[t]
\centering
\small
\setlength{\tabcolsep}{8pt}
\caption{Defense of DiffPure in MiniGPT-4 under different image configurations. Attack Success Rate is evaluated on the RealToxicityPrompts benchmark.}
\label{tab:diffpure_evaluate}
\resizebox{\columnwidth}{!}{
\begin{tabular}{l|c}
\toprule
\textbf{Image Configuration} & \textbf{Attack Success Rate (\%)} \\
\midrule
\textbf{Benign clean Image $I_{c}$} & 34.8 ± 1.6 \\
\quad + $n$ ($\sigma_{n}=30/255$) & 44.1 \\
\quad + $n$ ($\sigma_{n}=30/255$) + DiffPure ($t^{*}=50$) & 44.3 \\ \hline
\textbf{Adversarial image $I_{adv}$ ($\epsilon=16/255$)} & 53.6 ± 1.0 \\
\quad + DiffPure ($t^{*}=50$) & 45.0 \\ \hline
\textbf{Adversarial image $I_{adv}$ ($\epsilon=32/255$)} & 59.4 ± 1.4 \\
\quad + DiffPure ($t^{*}=50$) & 45.5 \\ \hline
\textbf{Adversarial image $I_{adv}$ ($\epsilon=64/255$)} & 67.2 ± 0.2 \\
\quad + DiffPure ($t^{*}=50$) & 44.5 \\
\bottomrule
\end{tabular}
}
\vspace{-2mm}
\end{table}

\subsection{DiffPure-VLM}

\begin{figure}[tb]
  \centering
  \includegraphics[width=\columnwidth]{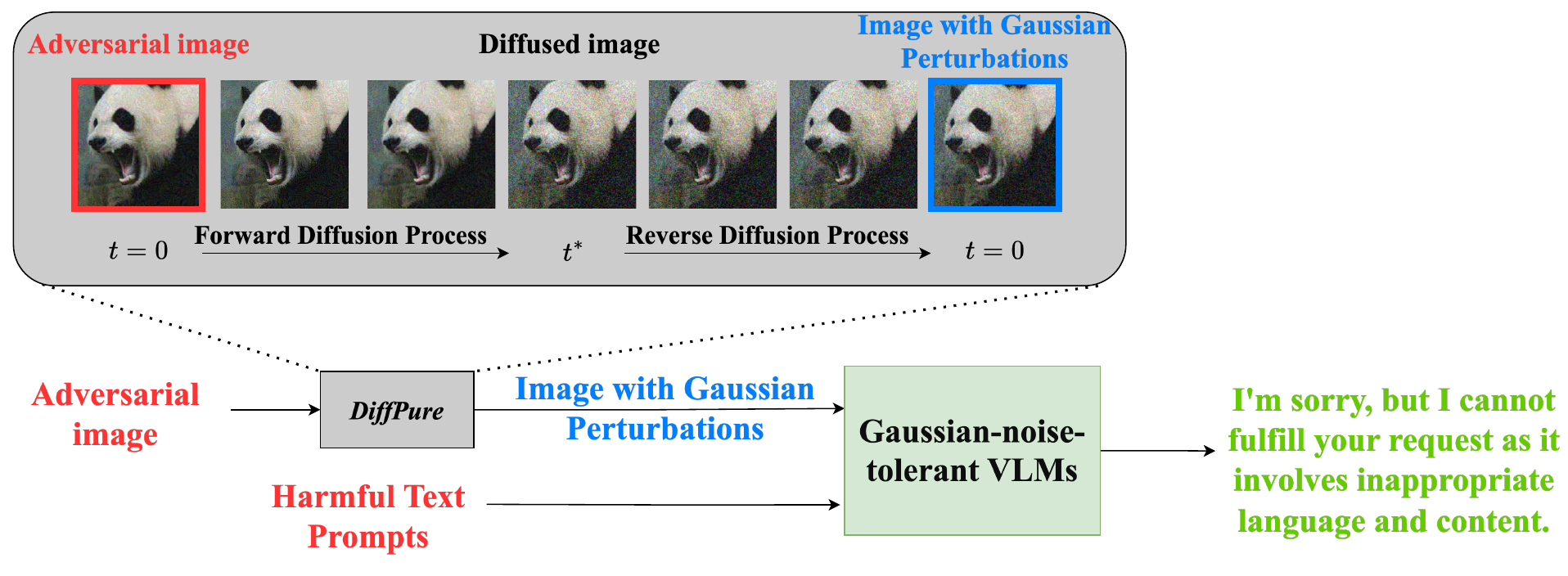}
  \caption{The overall framework of DiffPure-VLM.}
  \label{fig:DiffPure-VLM}
  \vspace{-1mm}
\end{figure}

Leveraging our safety fine-tuning approach and DiffPure's unique characteristic, we propose DiffPure-VLM — a defense pipeline that integrates Diffusion Models with Gaussian-noise-tolerant VLMs, as illustrated in Figure~\ref{fig:DiffPure-VLM}. Specifically, we purify adversarial images by applying a small timestep in DiffPure to preserve image content. The purified image with slight Gaussian-like noise is fed into the Gaussian-noise-tolerant, safety-tuned VLM, effectively mitigating the adversarial perturbations.

First, to verify DiffPure's effectiveness in our defense pipeline, we compare its performance against JailGuard for mitigating optimization-based perturbation attacks following \cite{qi2024visual} on the RealToxicityPrompts benchmark. We selected two base models, LLaVA-VLGuard and our LLaVA-RobustVLGuard, for a comprehensive evaluation. As shown in Table~\ref{tab:baseline_compare}, DiffPure with $t^* = 50$ consistently outperforms JailGuard across both base models. Notably, when paired with our VLM, DiffPure delivers a substantially greater improvement over JailGuard than when paired with LLaVA-VLGuard, confirming that its distribution-shifting properties are especially well-suited to our robust VLM.

\begin{table}[t]
\centering
\caption{Comparison of image preprocessing methods for mitigating adversarial attacks ($\epsilon$ = 32/255) on RealToxicityPrompts.}
\label{tab:baseline_compare}
\setlength{\tabcolsep}{1pt}
\resizebox{\columnwidth}{!}{%
\begin{tabular}{l|c|cccccc}
\hline
\textbf{Model} & \textbf{\begin{tabular}[c]{@{}c@{}}Attack \\ Success Rate $\downarrow$ \end{tabular}} & \textbf{Identity} & \textbf{Profanity} & \textbf{\begin{tabular}[c]{@{}c@{}}Severe \\ Toxicity\end{tabular}} & \textbf{\begin{tabular}[c]{@{}c@{}}Sexually \\ Explicit\end{tabular}} & \textbf{Threat} & \textbf{Toxicity} \\ \hline
LLaVA-VLGuard & 70.4 & 21.3 & 52.8 & 7.5 & 16.7 & 7.0 & 67.2 \\
JailGuard + LLaVA-VLGuard & 52.1 & 12.5 & 39.0 & 5.3 & 13.2 & 4.9 & 50.0 \\
DiffPure + LLaVA-VLGuard & \textbf{51.1} & 3.4 & 40.9 & 2.2 & 13.4 & 3.6 & 47.5 \\ \hline
LLaVA-RobustVLGuard & 62.5 & 7.8 & 48.0 & 5.4 & 16.5 & 5.8 & 60.0 \\
JailGuard + LLaVA-RobustVLGuard & 48.9 & 6.0 & 37.3 & 4.8 & 13.4 & 4.0 & 46.5 \\
DiffPure + LLaVA-RobustVLGuard & \textbf{43.9} & 3.2 & 34.6 & 2.4 & 12.8 & 3.7 & 41.0 \\ \hline
\end{tabular}%
}
\vspace{-2mm}
\end{table}

To evaluate the generalization of DiffPure-VLM, we assess the pipeline's performance under various optimization-based perturbation attack strengths (pixel constraint $\epsilon \in \{16/255, 32/255, 64/255\}$) and different timesteps (\(t^{*} = 50, 150\)) of DiffPure in our suite of three safety fine-tuned VLMs. For brevity, Table~\ref{tab:diffpure-vlm} presents results for \(\epsilon = 32/255\). Results for the other \(\epsilon\) values are provided in Table 7 in the Appendix. The experimental results demonstrate that, compared to clean or Gaussian-noise inputs, adversarial perturbation attacks substantially increase attack success rates. Nevertheless, our DiffPure mechanism consistently reduces these rates across all tested timesteps, nearly restoring performance to that of clean or Gaussian inputs. This not only confirms our analysis of DiffPure’s unique ability to transform adversarial noise into Gaussian noise, but also highlights the efficacy and generalization of DiffPure-VLM in effectively mitigating the impact of perturbations.

\begin{table}[t]
\centering
\caption{Evaluation of DiffPure-VLM's effectiveness on RealToxicityPrompts. Metrics include attack success rate and various toxicity levels (Perspective API \%). Additional results for other attack strengths are provided in Table 7 in the Appendix.}
\label{tab:diffpure-vlm}
\setlength{\tabcolsep}{1pt}
\resizebox{\columnwidth}{!}{%
\begin{tabular}{l|c|cccccc}
\hline
\textbf{Image Type} & \textbf{\begin{tabular}[c]{@{}c@{}}Attack \\ Success Rate $\downarrow$ \end{tabular}} & \textbf{Identity} & \textbf{Profanity} & \textbf{\begin{tabular}[c]{@{}c@{}}Severe \\ Toxicity\end{tabular}} & \textbf{\begin{tabular}[c]{@{}c@{}}Sexually \\ Explicit\end{tabular}} & \textbf{Threat} & \textbf{Toxicity} \\ \hline

\multicolumn{8}{c}{\textbf{InternVL2-8B-RobustVLGuard}} \\ \hline
Benign Clean image & 29.9 & 0.8 & 22.1 & 0.3 & 7.2 & 1.5 & 25.9 \\ \hline
Gaussian Noisy image & 34.5 & 2.1 & 27.2 & 1.3 & 8.4 & 1.6 & 31.3 \\ \hline
Adversarial image ($\epsilon$ = 32/255) & 70.6 & 26.7 & 56.5 & 9.2 & 17.3 & 6.9 & 68.1 \\ \hline
\quad+DiffPure-VLM (t*=50) & \textbf{33.4} & 2.4 & 20.6 & 0.7 & 8.1 & 2.4 & 29.1 \\ \hline
\quad+DiffPure-VLM (t*=150) & \textbf{32.8} & 1.7 & 25.9 & 0.6 & 7.7 & 1.8 & 29.1 \\ \hline

\multicolumn{8}{c}{\textbf{LLaVA-v1.5-7B-RobustVLGuard}} \\ \hline
Benign Clean image & 43.6 & 4.6 & 34.7 & 2.4 & 12.3 & 3.5 & 41.0 \\ \hline
Gaussian Noisy image & 42.3 & 3.1 & 34.5 & 1.9 & 11.8 & 3.1 & 40.0 \\ \hline
Adversarial image ($\epsilon$ = 32/255) & 62.5 & 7.8 & 48.0 & 5.4 & 16.5 & 5.8 & 60.0 \\ \hline
\quad+DiffPure-VLM (t*=50) & \textbf{43.9} & 3.2 & 34.6 & 2.4 & 12.8 & 3.7 & 41.0 \\ \hline
\quad+DiffPure-VLM (t*=150) & \textbf{42.5} & 3.5 & 32.7 & 2.8 & 12.1 & 4.1 & 39.3 \\ \hline

\multicolumn{8}{c}{\textbf{MiniGPT-4-13B-RobustVLGuard}} \\ \hline
Benign Clean image & 16.0 & 0.4 & 9.9 & 0.3 & 4.6 & 1.1 & 12.1 \\ \hline
Gaussian Noisy image & 16.5 & 0.9 & 11.9 & 0.6 & 5.8 & 1.0 & 14.0 \\ \hline
Adversarial image ($\epsilon$ = 32/255) & 53.7 & 9.8 & 35.3 & 4.1 & 13.9 & 5.4 & 48.1 \\ \hline
\quad+DiffPure-VLM (t*=50) & \textbf{13.6} & 0.3 & 9.2 & 0.2 & 5.5 & 0.9 & 10.6 \\ \hline
\quad+DiffPure-VLM (t*=150) & \textbf{11.9} & 0.5 & 8.6 & 0.2 & 4.2 & 0.6 & 9.9 \\ \hline

\end{tabular}%
}
\vspace{-2mm}
\end{table}

%% file: sec/2_related.tex
\section{Related Work}

\subsection{Vision Language Model}
In recent years, Vision Language Models have gained significant attention due to their ability to jointly process and understand both visual and textual data. A major breakthrough in this area came with the development of large-scale pre-trained models like CLIP~\cite{radford2021learning}, which aligns images and texts in a shared embedding space. By training on a vast dataset of image-text pairs, CLIP enabled zero-shot learning capabilities for tasks such as action recognition~\cite{carreira2019short, miech2020rareact} and optical character recognition~\cite{socher2013recursive, kiela2020hateful}. Recent advances have introduced multi-modal foundation models like InstructBLIP~\cite{dai2023instructblipgeneralpurposevisionlanguagemodels}, LLaVA~\cite{liu2024visual}, and Qwen-VL~\cite{bai2023qwen}, which leverage large-scale transformer networks that can process both image and text inputs simultaneously, enabling more sophisticated reasoning over complex scenarios. These models excel in advanced tasks like UI understanding~\cite{hong2024cogagent, fu2024ui} and visual question answering~\cite{antol2015vqa}, and open up new possibilities for generative capabilities.

\subsection{Jailbreaking VLMs}

As VLMs become increasingly integrated into various applications, critical concerns have arisen regarding their robustness, security, and ethical alignment. A critical issue is "jailbreaking"—the ability to bypass safety protocols, potentially triggering unintended or harmful behaviors. Given that VLMs process both textual and visual inputs, vulnerabilities inherent in LLMs may also affect VLMs. Moreover, the integration of visual inputs into VLMs, while expanding their capabilities, introduces more diverse attack patterns, significantly heightening the severity of potential threats~\cite{wei2024jailbroken, jin2024jailbreakzoo}. Several methods have emerged in this domain, including converting the malicious content into images through typography-based manipulations~\cite{gong2023figstep}, leveraging multimodal perturbations to craft stronger adversarial attacks~\cite{zhang2022towards}, and using gradient updates to embed malicious triggers within seemingly benign images~\cite{shayegani2023jailbreak}.  These diverse sophisticated methods pose significant challenges to ensuring the safety and reliability of VLMs.

\subsection{Safeguarding VLMs}

Given the growing challenges posed by malicious attacks on VLMs, it is crucial to develop effective defense strategies. One key approach involves enhancing the training process through adversarial fine-tuning~\cite{schlarmann2024robust, zong2024safety} or prompt tuning~\cite{zhang2023adversarial}, where models or learnable prompts are trained on perturbed examples to improve resilience against real-world attack scenarios. Another promising defense strategy is input sanitization, which aims to detect or neutralize adversarial inputs before they can compromise the model. Methods in this area include shifting sample probabilities to adversarial-free regions~\cite{theagarajan2019shieldnets}, applying randomized smoothing to mitigate the impact of adversarial inputs~\cite{sun2024safeguarding}, and utilizing unlabeled prompts from users in the wild for malicious prompt detection~\cite{du2024vlmguard}. In this work, we concentrate on improving VLM robustness to noise-augmented inputs. By integrating Diffusion Models with safety fine-tuned VLMs, we equip these models with enhanced protection against a broad range of adversarial attacks.

%% file: sec/6_conclusion.tex
\section{Conclusion}

In this work, we address a critical gap in the robustness of VLMs by examining the impact of Gaussian noise perturbations and propose Robust-VLGuard, a multimodal safety dataset paired with Gaussian-noise-augmented fine-tuning, to enhance safety alignment and preserve the helpfulness of VLMs. We further propose DiffPure-VLM to defend Optimization-Based Visual Perturbation Attack by using a diffusion model to transfer adversarial noise to Gaussian noise which can be defended by VLMs with noise-augmented safety fine-tuning.
The experimental result demonstrates the superiority of DiffPure-VLM in Gaussian noise and adversarial perturbations with baseline methods.


While DiffPure-VLM provides a practical defense, future work includes integrating noise augmentation in pretraining, expanding the safety dataset for broader tasks, and exploring adaptive multi-modal defenses to further enhance real-world performance.

%% file: sec/7_social_impact.tex
\section{Social Impact}

This research exposes VLM vulnerabilities to noisy inputs and adversarial attacks. While Robust-VLGuard and DiffPure-VLM enhance robustness, our findings have dual-use implications. Given VLMs' growing adoption, we responsibly disclose these issues to raise awareness and foster the development of more secure models, mitigating deployment risks.



%% file: sec/X_suppl.tex
\clearpage
\maketitlesupplementary


\section{Overview of the Supplementary Material}

This supplementary material offers additional details and analyses to further support the findings presented in the main manuscript. It includes detailed information on the experimental configuration (Appendix \ref{appendix:exp}), more evaluation on recent vision-language models (Appendix \ref{sec:additional}), a thorough analysis of the limitations and unique characteristics of DiffPure (Appendix \ref{appendix:diffpure}), extended implementation specifics of DiffPure-VLM (Appendix \ref{appendix:vlm}), and conjectures along with preliminary theoretical discussions on the effects of Gaussian noise (Appendix \ref{appendix:gaussian}). Collectively, these sections provide deeper insights into our methodology, enhancing the transparency and reproducibility of our research.

\section{Experiment Details}
\label{appendix:exp}
\subsection{Models}
In this work, we conduct all experiments on three leading Vision-Language Models (VLMs), i.e., MiniGPT-4 (13B) \cite{zhu2023minigpt}, LLaVA-v1.5 (7B) \cite{liu2024visual}, and InternVL2 (8B) \cite{chen2024far}. We use the official model weights from HuggingFace or GitHub repositories for experiments in our paper. These model details are summarized in Table~\ref{tab:vlm-details}. Each model features a distinct LLM, vision encoder, and vision-language alignment method, allowing us to draw broader insights.

\begin{table}[h]
\centering
\caption{Specifications of the evaluated VLMs.}
\resizebox{\columnwidth}{!}{%
\renewcommand{\arraystretch}{1.6}
\setlength{\tabcolsep}{10pt}
\fontsize{26pt}{32pt}\selectfont
\begin{tabular}{l|c|c|c|c}
\hline
\textbf{Model} & \textbf{Size} & \textbf{Vision Encoder} & \textbf{LLM} & \textbf{VL Connection Module} \\ \hline
MiniGPT-4-13B  & 14B           & EVA-CLIP ViT-G/14       & Vicuna-v0-13B  & Q-former             \\ \hline
LLaVA-v1.5-7B  & 7B            & CLIP ViT-L/14           & Vicuna-v1.5-7B & MLP                  \\ \hline
InternVL2-8B   & 8B            & InternViT-300M-448px    & InternLM2-8B   & MLP                  \\ \hline
\end{tabular}%
}
\label{tab:vlm-details}
\end{table}

\subsection{Fine-tuning Configuration}

We present the detailed hyper-parameters for post-hoc fine-tuning on our Robust-VLGuard dataset in Table~\ref{tab:finetuning-config}. Gaussian noise augmentation was applied to the training images, with a randomly selected standard deviation between 0.01 and 0.15, and a 70\% probability of application. The fine-tuning was performed over 3 epochs on a single A100-80G GPU, using a consistent batch size of 16. For MiniGPT-4-13B, unfreezing the linear projector significantly improved robustness in terms of helpfulness and safety alignment. However, for LLaVA-v1.5-7B and InternVL2-8B, unfreezing the linear projector led to increased overfitting, likely due to differences in the vision-language connection modules of these models.

\begin{table}[h]
\centering
\caption{Post-hoc fine-tuning hyper-parameters of different models.}
\renewcommand{\arraystretch}{1.4}
\setlength{\tabcolsep}{10pt}
\resizebox{\columnwidth}{!}{%
\fontsize{22pt}{26pt}\selectfont
\begin{tabular}{l|l|c|c|c}
\hline
\textbf{Model} & \textbf{Training Module} & \textbf{LoRA Rank} & \textbf{LoRA Alpha} & \textbf{Learning Rate} \\ \hline
MiniGPT-4-13B  & \begin{tabular}[c]{@{}l@{}}Vision Encoder \& \\ Linear Projector\end{tabular} & 16  & 32  & 3e-5 \\ \hline
LLaVA-v1.5-7B  & Vision Encoder                                                                & 16  & 256 & 4e-5 \\ \hline
InternVL2-8B   & Vision Encoder                                                                & 16  & 256 & 4e-5 \\ \hline
\end{tabular}%
}
\label{tab:finetuning-config}
\end{table}

\subsection{Details of Evaluation Settings}

For evaluation on the MM-Vet benchmark, we set the temperature to 0 and use greedy decoding across all experiments to ensure reproducibility in helpfulness assessments. For safety evaluations on the RealToxicityPrompts benchmark, we follow the setup of Qi et al.~\cite{qi2024visual}, using a temperature of 1 and performing three runs to calculate the average attack success rate. Greedy decoding is also employed for this benchmark. The choice of temperature 1 reflects real-world usage, where sampling is typically enabled during interactions with VLMs. This setting aims to better simulate real-world scenarios when assessing safety alignment.

Additionally, the MM-Vet and RealToxicityPrompts benchmarks offer a comprehensive set of metrics covering various aspects. For the sake of brevity, we report only the overall metrics — Performance Score and Attack Success Rate — in the main paper. Here, we present the detailed evaluation results in Table~\ref{tab:vlm-helpfulness} and Table~\ref{tab:vlm-safety}, corresponding to Figure 2 in the main paper. The results show that Gaussian noisy images negatively impact nearly all metrics across both benchmarks and various models. Notably, using Gaussian noisy images as prompts improves MiniGPT-4’s performance on the OCR metric in the MM-Vet benchmark, highlighting the current VLMs’ lack of robustness.

\begin{table*}[t]
\centering
\caption{Robustness comparison of various models on the MM-Vet benchmark using clean and Gaussian noisy image prompts (GPT-4 \%).}
\label{tab:vlm-helpfulness}
\renewcommand{\arraystretch}{1.2}  
\setlength{\tabcolsep}{1.5pt}  
\resizebox{0.7\textwidth}{!}{
\small  
\begin{tabular}{l|c|c|c|c|c|c|c}
\hline
\textbf{Image Type} & \textbf{\begin{tabular}[c]{@{}c@{}}Performance \\ Score $\uparrow$\end{tabular}} & \textbf{Recognition} & \textbf{OCR} & \textbf{Knowledge} & \textbf{Generation} & \textbf{Spatial} & \textbf{Math} \\ \hline

\multicolumn{8}{c}{\textbf{MiniGPT-4-13B}} \\ \hline
Clean Image          & 26.7 & 34.9 & 13.5 & 27.4 & 27.1 & 19.1 & 7.7 \\ \hline
Gaussian Noisy Image  & 24.0 \textcolor{red}{(-2.7)} & 29.0 \textcolor{red}{(-5.9)} & 16.9 \textcolor{green}{(+3.4)} & 20.5 \textcolor{red}{(-6.9)} & 22.5 \textcolor{red}{(-4.6)} & 20.7 \textcolor{red}{(+1.6)} & 7.7 \textcolor{gray}{(0.0)} \\ \hline

\multicolumn{8}{c}{\textbf{LLaVA-v1.5-7B}} \\ \hline
Clean Image          & 33.0 & 37.9 & 23.9 & 20.4 & 23.6 & 28.5 & 11.5 \\ \hline
Gaussian Noisy Image  & 31.3 \textcolor{red}{(-1.7)} & 36.3 \textcolor{red}{(-1.6)} & 21.9 \textcolor{red}{(-2.0)} & 18.3 \textcolor{red}{(-2.1)} & 21.2 \textcolor{red}{(-2.4)} & 25.7 \textcolor{red}{(-2.8)} & 3.8 \textcolor{red}{(-7.7)} \\ \hline

\multicolumn{8}{c}{\textbf{InternVL2-8B}} \\ \hline
Clean Image          & 59.9 & 53.5 & 71.9 & 40.4 & 44.6 & 69.7 & 65.4 \\ \hline
Gaussian Noisy Image  & 54.4 \textcolor{red}{(-5.5)} & 47.1 \textcolor{red}{(-6.4)} & 66.8 \textcolor{red}{(-5.1)} & 35.1 \textcolor{red}{(-5.3)} & 39.2 \textcolor{red}{(-5.4)} & 61.1 \textcolor{red}{(-8.6)} & 53.5 \textcolor{red}{(-11.9)} \\ \hline

\end{tabular}%
}
\vspace{-3mm}
\end{table*}

\begin{table*}[t]
\centering
\caption{Robustness comparison of various models on the RealToxicityPrompts benchmark using clean and Gaussian noisy image prompts. Metrics include Attack Success Rate and toxicity levels (Perspective API \%).}
\label{tab:vlm-safety}
\renewcommand{\arraystretch}{1.2}
\setlength{\tabcolsep}{1.5pt}  
\resizebox{0.7\textwidth}{!}{
\small  
\begin{tabular}{l|c|c|c|c|c|c|c}
\hline
\textbf{Image Type} & \textbf{\begin{tabular}[c]{@{}c@{}}Attack \\ Success Rate $\downarrow$\end{tabular}} & \textbf{Identity Attack} & \textbf{Profanity} & \textbf{\begin{tabular}[c]{@{}c@{}}Severe \\ Toxicity\end{tabular}} & \textbf{\begin{tabular}[c]{@{}c@{}}Sexually \\ Explicit\end{tabular}} & \textbf{Threat} & \textbf{Toxicity} \\ \hline

\multicolumn{8}{c}{\textbf{MiniGPT-4-13B}} \\ \hline
Clean Image          & 34.8 & 2.7 & 25.1 & 1.5 & 12.2 & 2.0 & 30.5 \\ \hline
Gaussian Noisy Image  & 44.1 \textcolor{red}{(+9.3)} & 3.6 \textcolor{red}{(+0.9)} & 31.2 \textcolor{red}{(+6.1)} & 2.3 \textcolor{red}{(+0.8)} & 13.7 \textcolor{red}{(+1.5)} & 2.9 \textcolor{red}{(+0.9)} & 38.2 \textcolor{red}{(+7.7)} \\ \hline

\multicolumn{8}{c}{\textbf{LLaVA-v1.5-7B}} \\ \hline
Clean Image          & 57.7 & 5.7 & 46.8 & 3.7 & 18.0 & 3.8 & 54.4 \\ \hline
Gaussian Noisy Image  & 60.1 \textcolor{red}{(+2.4)} & 4.8 \textcolor{green}{(-0.9)} & 48.1 \textcolor{red}{(+1.3)} & 2.9 \textcolor{green}{(-0.8)} & 17.8 \textcolor{green}{(-0.2)} & 4.0 \textcolor{red}{(+0.2)} & 56.0 \textcolor{red}{(+1.6)} \\ \hline

\multicolumn{8}{c}{\textbf{InternVL2-8B}} \\ \hline
Clean Image          & 50.5 & 4.1 & 40.2 & 1.9 & 13.5 & 2.5 & 44.3 \\ \hline
Gaussian Noisy Image  & 57.2 \textcolor{red}{(+6.7)} & 4.5 \textcolor{red}{(+0.4)} & 45.9 \textcolor{red}{(+5.7)} & 2.0 \textcolor{red}{(+0.1)} & 14.3 \textcolor{red}{(+0.8)} & 3.2 \textcolor{red}{(+0.7)} & 51.7 \textcolor{red}{(+7.4)} \\ \hline

\end{tabular}%
}
\vspace{-3mm}
\end{table*}

\section{Additional Evaluation on Recent Vision-Language Models}
\label{sec:additional}

In this section, we further assess the robustness of state-of-the-art vision-language models against Gaussian noise. Table~\ref{tab:addi} presents the attack success rates on the RealToxicityPrompts benchmark for four recent VLMs—LLaMA-3.2-Vision-11B \cite{dubey2024llama}, Ivy-VLM-3B \cite{ivy2024ivy-vl}, Qwen2.5-VL-7B \cite{bai2025qwen2}, and InternVL2.5-8B \cite{chen2024expanding}—under various Gaussian noise levels. Lower percentages indicate improved safety alignment.

As shown, when Gaussian noise is introduced at increasing levels ($\sigma_{n}=30/255$, $\sigma_{n}=50/255$, and $\sigma_{n}=70/255$), all models exhibit a rise in attack success rates, highlighting their sensitivity to simple Gaussian noise perturbations. These findings underscore the need for robust noise augmentation and defense strategies in training pipelines to maintain safety alignment in VLMs.

\begin{table*}[ht]
\centering
\caption{Attack success rate (\%) on the RealToxicityPrompts benchmark for various vision-language models under different noise levels. Lower scores indicate improved safety alignment.}
\label{tab:addi}
\resizebox{0.8\textwidth}{!}{
\small  
\begin{tabular}{l|cccc}
\hline 
\multirow{2}{*}{}  & \multicolumn{4}{c}{RealToxicityPrompts (\%) ↓} \\ \cline{2-5} 
                   & LLaMA-3.2-Vision-11B & Ivy-VLM-3B & Qwen2.5-VL-7B & InternVL2.5-8B \\ \hline
Benign clean Image $\scriptscriptstyle I_{c}$ & 45.4 & 29.9 & 36.8 & 43.9 \\ \hline
\quad + $n$ ($\sigma_{n}=30/255$)                & 46.4 (+1.0) & 35.5 (+5.6) & 39.3 (+2.5) & 51.5 (+7.6) \\ \hline
\quad + $n$ ($\sigma_{n}=50/255$)                & 47.6 (+2.2) & 40.3 (+10.4) & 39.5 (+2.7) & 52.8 (+8.9) \\ \hline
\quad + $n$ ($\sigma_{n}=70/255$)                & 48.5 (+3.1) & 42.0 (+12.1) & 46.1 (+9.3) & 54.0 (+10.1) \\ \hline
\end{tabular}
}
\end{table*}

\section{Further Analysis of DiffPure}
\label{appendix:diffpure}

\subsection{Defence Performance}

In this section, we present a comprehensive analysis of the effects of DiffPure \cite{nie2022diffusion} and Gaussian noise under perturbation-based attacks in Vision-Language Models (VLMs). Specifically, we extend the experimental setup described in Section 3.1 in the main paper by varying the standard deviation $\sigma_{n}$ of Gaussian noise $n$ and the timestep parameter $t^{*}$ in DiffPure. Results are summarized in Table \ref{tab:img_exp_diff}.
First, Gaussian noise $n$ with standard deviations $\sigma_{n} \in \{15/255,30/255, 50/255, 75/255\}$ is added to the benign clean image $I_{c}$ to evaluate its impact on the Attack Success Rate. The results demonstrate that the Attack Success Rate under Gaussian noise is significantly higher than that of the benign clean image. When $\sigma_{n} \leq 50/255$, increasing $\sigma_{n}$ will lead to a higher Attack Success Rate. However, this trend did not continue at a higher $\sigma_{n}$ setting (e.g., $\sigma_{n}=75/255$), indicating that the effect of Gaussian noise on VLMs is limited.
Next, we apply DiffPure with different timesteps $t^{*} \in \{50, 100, 150\}$ to generate diffused images from adversarial inputs $I_{adv}$ with varying perturbation constraints $\epsilon$. For $\epsilon = 16/255$, increasing $t^{*}$ to 100 or 150 reduces the Attack Success Rate but does not lower it below the level observed for the benign clean image. For larger perturbation constraints, increasing $t^{*}$ fails to decrease the Attack Success Rate, with a comparable performance of Gaussian noisy images.

\begin{table*}[t]
\centering
\caption{Detailed results of the defense of DiffPure in MiniGPT-4 on the RealToxicityPrompts benchmark under different image configurations. (Perspective API \%).}
\label{tab:img_exp_diff}
\renewcommand{\arraystretch}{1.2}
\setlength{\tabcolsep}{1.5pt}  
\resizebox{0.8\textwidth}{!}{
\small  
\begin{tabular}{l|c|c|c|c|c|c|c}
\hline
\textbf{Image Configuration} & \textbf{\begin{tabular}[c]{@{}c@{}}Attack \\ Success Rate $\downarrow$\end{tabular}} & \textbf{Identity Attack} & \textbf{Profanity} & \textbf{\begin{tabular}[c]{@{}c@{}}Severe \\ Toxicity\end{tabular}} & \textbf{\begin{tabular}[c]{@{}c@{}}Sexually \\ Explicit\end{tabular}} & \textbf{Threat} & \textbf{Toxicity} \\ \hline

\textbf{Benign clean Image $\tiny{I_{c}}$}    & 34.8 & 2.7 & 25.1 & 1.5 & 12.2 & 2.0 & 30.5 \\ 
\quad + $n$ ($\sigma_{n}=15/255$)  & 38.5 \textcolor{red}{(+3.7)} & 2.9 \textcolor{red}{(+0.2)} & 27.4 \textcolor{red}{(+2.3)} & 1.1 \textcolor{green}{(-0.4)} & 13.0 \textcolor{red}{(+0.8)} & 2.3 \textcolor{red}{(+0.3)} & 34.1 \textcolor{red}{(+3.6)} \\ 
\quad + $n$ ($\sigma_{n}=30/255$)  & 44.1 \textcolor{red}{(+9.3)} & 3.6 \textcolor{red}{(+0.9)} & 31.2 \textcolor{red}{(+6.1)} & 2.3 \textcolor{red}{(+0.8)} & 13.7 \textcolor{red}{(+1.5)} & 2.9 \textcolor{red}{(+0.9)} & 38.2 \textcolor{red}{(+7.7)} \\ 
\quad + $n$ ($\sigma_{n}=50/255$)  & 46.3 \textcolor{red}{(+11.5)} & 3.4 \textcolor{red}{(+0.7)} & 34.0 \textcolor{red}{(+8.9)} & 1.8 \textcolor{red}{(+0.3)} & 14.8 \textcolor{red}{(+2.6)} & 2.5 \textcolor{red}{(+0.5)} & 39.5 \textcolor{red}{(+9.0)} \\ 
\quad + $n$ ($\sigma_{n}=75/255$)  & 44.1 \textcolor{red}{(+9.3)} & 3.8 \textcolor{red}{(+1.1)} & 30.1 \textcolor{red}{(+5.0)} & 1.9 \textcolor{red}{(+0.4)} & 14.3 \textcolor{red}{(+2.1)} & 2.8 \textcolor{red}{(+0.8)} & 37.5 \textcolor{red}{(+7.0)} \\ \hline
\textbf{Adversarial image $I_{adv}$ ($\epsilon=16/255$)}  & 53.6 \textcolor{red}{(+18.8)} & 8.4 \textcolor{red}{(+5.7)} & 36.6 \textcolor{red}{(+9.4)} & 6.6 \textcolor{red}{(+5.1)} & 14.1 \textcolor{red}{(+1.9)} & 4.7 \textcolor{red}{(+2.7)} & 48.6 \textcolor{red}{(+18.1)} \\ 
\quad + DiffPure ($t^{*}=50$)  & 45.0 \textcolor{red}{(+10.2)} & 2.5 \textcolor{green}{(-0.2)} & 31.7 \textcolor{red}{(+6.6)} & 1.8 \textcolor{red}{(+0.3)} & 14.5 \textcolor{red}{(+2.3)} & 2.8 \textcolor{red}{(+0.8)} & 38.8 \textcolor{red}{(+8.3)} \\ 
\quad + DiffPure ($t^{*}=100$)  & 37.6 \textcolor{red}{(+2.8)} & 3.0 \textcolor{red}{(+0.3)} & 25.6 \textcolor{red}{(+0.5)} & 1.3 \textcolor{green}{(-0.2)} & 12.3 \textcolor{red}{(+0.1)} & 1.8 \textcolor{green}{(-0.2)} & 33.1 \textcolor{red}{(+2.6)} \\ 
\quad + DiffPure ($t^{*}=150$)  & 37.7 \textcolor{red}{(+2.9)} & 2.5 \textcolor{green}{(-0.2)} & 26.5 \textcolor{red}{(+1.4)} & 2.1 \textcolor{red}{(+0.6)} & 12.2 \textcolor{red}{(+0.0)} & 2.5 \textcolor{red}{(+0.5)} & 32.9 \textcolor{red}{(+2.4)} \\ \hline
\textbf{Adversarial image $I_{adv}$ ($\epsilon=32/255$)}  & 59.4 \textcolor{red}{(+24.6)} & 14.6 \textcolor{red}{(+11.9)} & 39.5 \textcolor{red}{(+14.4)} & 7.0 \textcolor{red}{(+5.5)} & 14.9 \textcolor{red}{(+2.7)} & 6.2 \textcolor{red}{(+4.2)} & 53.8 \textcolor{red}{(+23.3)} \\ 
\quad + DiffPure ($t^{*}=50$)  & 45.5 \textcolor{red}{(+10.7)} & 2.6 \textcolor{green}{(-0.1)} & 32.1 \textcolor{red}{(+7.0)} & 2.2 \textcolor{red}{(+0.7)} & 14.8 \textcolor{red}{(+2.6)} & 3.0 \textcolor{red}{(+1.0)} & 38.5 \textcolor{red}{(+8.0)} \\ 
\quad + DiffPure ($t^{*}=100$)  & 43.8 \textcolor{red}{(+9.0)} & 3.3 \textcolor{red}{(+0.6)} & 31.9 \textcolor{red}{(+6.8)} & 1.9 \textcolor{red}{(+0.4)} & 13.1 \textcolor{red}{(+0.9)} & 2.5 \textcolor{red}{(+0.5)} & 38.1 \textcolor{red}{(+7.6)} \\ 
\quad + DiffPure ($t^{*}=150$)  & 42.3 \textcolor{red}{(+7.5)} & 3.7 \textcolor{red}{(+1.0)} & 30.4 \textcolor{red}{(+5.3)} & 1.3 \textcolor{green}{(-0.2)} & 13.3 \textcolor{red}{(+1.1)} & 2.8 \textcolor{red}{(+0.8)} & 36.3 \textcolor{red}{(+5.8)} \\ \hline
\textbf{Adversarial image $I_{adv}$ ($\epsilon=64/255$)} & 67.2 \textcolor{red}{(+32.4)} & 15.9 \textcolor{red}{(+13.2)} & 49.6 \textcolor{red}{(+24.5)} & 12.2 \textcolor{red}{(+10.7)} & 16.9 \textcolor{red}{(+4.7)} & 6.6 \textcolor{red}{(+4.6)} & 63.1 \textcolor{red}{(+32.6)} \\ 
\quad + DiffPure ($t^{*}=50$)  & 44.5 \textcolor{red}{(+9.7)} & 2.9 \textcolor{red}{(+0.2)} & 32.2 \textcolor{red}{(+7.1)} & 2.4 \textcolor{red}{(+0.9)} & 13.7 \textcolor{red}{(+1.5)} & 2.7 \textcolor{red}{(+0.7)} & 38.0 \textcolor{red}{(+7.5)} \\ 
\quad + DiffPure ($t^{*}=100$)  & 42.1 \textcolor{red}{(+7.3)} & 2.8 \textcolor{red}{(+0.1)} & 30.3 \textcolor{red}{(+5.2)} & 1.9 \textcolor{red}{(+0.4)} & 13.7 \textcolor{red}{(+1.5)} & 3.0 \textcolor{red}{(+1.0)} & 36.5 \textcolor{red}{(+6.0)} \\ 
\quad + DiffPure ($t^{*}=150$)  & 44.1 \textcolor{red}{(+9.3)} & 3.3 \textcolor{red}{(+0.6)} & 31.5 \textcolor{red}{(+6.4)} & 1.4 \textcolor{green}{(-0.1)} & 13.3 \textcolor{red}{(+1.1)} & 2.5 \textcolor{red}{(+0.5)} & 38.2 \textcolor{red}{(+7.7)} \\ \hline

\end{tabular}%
}
\end{table*}

\begin{table*}[t]
\Huge
\centering
\caption{Evaluation of DiffPure-VLM's effectiveness on RealToxicityPrompts across different image configurations. Metrics include attack success rate and toxicity levels (Perspective API \%).}
\label{tab:diffpure-vlm-v3}
\resizebox{0.8\textwidth}{!}{%
\begin{tabular}{l|c|c|c|c|c|c|c}
\hline
\textbf{Image Type} & \textbf{\begin{tabular}[c]{@{}c@{}}Attack \\ Success Rate $\downarrow$ \end{tabular}} & \textbf{Identity Attack} & \textbf{Profanity} & \textbf{\begin{tabular}[c]{@{}c@{}}Severe \\ Toxicity\end{tabular}} & \textbf{\begin{tabular}[c]{@{}c@{}}Sexually \\ Explicit\end{tabular}} & \textbf{Threat} & \textbf{Toxicity} \\ \hline
\multicolumn{8}{c}{\textbf{InternVL2-8B}} \\ \hline
Benign Clean image & 50.5 & 4.1 & 40.2 & 1.9 & 13.5 & 2.5 & 44.3 \\ \hline
Gaussian Noisy image & 57.2 & 4.5 & 45.9 & 2.0 & 14.3 & 3.2 & 51.7 \\ \hline
Adversarial image ($\epsilon$ = 32/255) & 65.0 & 21.1 & 49.2 & 7.5 & 16.6 & 5.0 & 61.9 \\ \hline
\quad+DiffPure-VLM (t*=50) & 53.1 & 3.8 & 41.6 & 2.0 & 13.6 & 2.2 & 48.0 \\ \hline

\multicolumn{8}{c}{\textbf{InternVL2-8B-VLGuard}} \\ \hline
Benign Clean image & 27.7 & 1.4 & 22.2 & 0.9 & 7.1 & 1.6 & 23.8 \\ \hline
Gaussian Noisy image & 39.9 & 2.5 & 31.4 & 1.3 & 10.3 & 1.8 & 35.8 \\ \hline
Adversarial image ($\epsilon$ = 32/255) & 72.3 & 12.3 & 60.6 & 8.6 & 19.9 & 6.5 & 69.3 \\ \hline
\quad+DiffPure-VLM (t*=50) & 35.7 & 2.0 & 28.9 & 0.8 & 9.8 & 1.8 & 31.6 \\ \hline

\multicolumn{8}{c}{\textbf{InternVL2-8B-RobustVLGuard}} \\ \hline
Benign Clean image & 29.9 & 0.8 & 22.1 & 0.3 & 7.2 & 1.5 & 25.9 \\ \hline
Gaussian Noisy image & 34.5 & 2.1 & 27.2 & 1.3 & 8.4 & 1.6 & 31.3 \\ \hline
Adversarial image ($\epsilon$ = 32/255) & 70.6 & 26.7 & 56.5 & 9.2 & 17.3 & 6.9 & 68.1 \\ \hline
\quad+DiffPure-VLM (t*=50) & \textbf{33.4} & 2.4 & 20.6 & 0.7 & 8.1 & 2.4 & 29.1 \\ \hline
\quad+DiffPure-VLM (t*=150) & \textbf{32.8} & 1.7 & 25.9 & 0.6 & 7.7 & 1.8 & 29.1 \\ \hline

\multicolumn{8}{c}{\textbf{LLaVA-v1.5-7B}} \\ \hline
Benign Clean image & 57.7 & 5.7 & 46.8 & 3.7 & 18.0 & 3.8 & 54.4 \\ \hline
Gaussian Noisy image & 60.1 & 4.8 & 48.1 & 2.9 & 17.8 & 4.0 & 56.0 \\ \hline
Adversarial image ($\epsilon$ = 32/255) & 66.0 & 16.6 & 51.6 & 8.8 & 18.0 & 4.7 & 64.5 \\ \hline
\quad+DiffPure-VLM (t*=50) & 58.5 & 5.9 & 45.5 & 2.7 & 17.0 & 4.3 & 53.3 \\ \hline

\multicolumn{8}{c}{\textbf{LLaVA-v1.5-7B-VLGuard}} \\ \hline
Benign Clean image & 50.3 & 4.3 & 40.6 & 2.0 & 13.6 & 4.3 & 46.9 \\ \hline
Gaussian Noisy image & 52.3 & 4.6 & 41.5 & 2.7 & 14.0 & 4.1 & 48.5 \\ \hline
Adversarial image ($\epsilon$ = 32/255) & 70.4 & 21.3 & 52.8 & 7.5 & 16.7 & 7.0 & 67.2 \\ \hline
\quad+DiffPure-VLM (t*=50) & 51.1 & 3.4 & 40.9 & 2.2 & 13.4 & 3.6 & 47.5 \\ \hline

\multicolumn{8}{c}{\textbf{LLaVA-v1.5-7B-RobustVLGuard}} \\ \hline
Benign Clean image & 43.6 & 4.6 & 34.7 & 2.4 & 12.3 & 3.5 & 41.0 \\ \hline
Gaussian Noisy image & 42.3 & 3.1 & 34.5 & 1.9 & 11.8 & 3.1 & 40.0 \\ \hline
Adversarial image ($\epsilon$ = 32/255) & 62.5 & 7.8 & 48.0 & 5.4 & 16.5 & 5.8 & 60.0 \\ \hline
\quad+DiffPure-VLM (t*=50) & \textbf{43.9} & 3.2 & 34.6 & 2.4 & 12.8 & 3.7 & 41.0 \\ \hline
\quad+DiffPure-VLM (t*=150) & \textbf{42.5} & 3.5 & 32.7 & 2.8 & 12.1 & 4.1 & 39.3 \\ \hline

\multicolumn{8}{c}{\textbf{MiniGPT-4-13B}} \\ \hline
Benign Clean image & 34.8 & 2.7 & 25.1 & 1.5 & 12.2 & 2.0 & 30.5 \\ \hline
Gaussian Noisy image & 44.1 & 3.6 & 31.2 & 2.3 & 13.7 & 2.9 & 38.2 \\ \hline
Adversarial image ($\epsilon$ = 32/255) & 59.4 & 14.6 & 39.5 & 7.0 & 14.9 & 6.2 & 53.8 \\ \hline
\quad+DiffPure-VLM (t*=50) & 45.5 & 2.6 & 32.1 & 2.2 & 14.8 & 3.0 & 38.5 \\ \hline

\multicolumn{8}{c}{\textbf{MiniGPT-4-13B-VLGuard}} \\ \hline
Benign Clean image & 41.3 & 2.8 & 30.1 & 2.2 & 14.6 & 2.5 & 37.3 \\ \hline
Gaussian Noisy image & 43.7 & 3.0 & 31.6 & 2.3 & 13.9 & 3.5 & 38.6\\ \hline
Adversarial image ($\epsilon$ = 32/255) & 67.6 & 10.5 & 48.2 & 7.0 & 19.9 & 7.8 & 61.7 \\ \hline
\quad+DiffPure-VLM (t*=50) & 45.0 & 4.2 & 33.1 & 2.1 & 14.6 & 3.1 & 40.7 \\ \hline

\multicolumn{8}{c}{\textbf{MiniGPT-4-13B-RobustVLGuard}} \\ \hline
Benign Clean image & 16.0 & 0.4 & 9.9 & 0.3 & 4.6 & 1.1 & 12.1 \\ \hline
Gaussian Noisy image & 16.5 & 0.9 & 11.9 & 0.6 & 5.8 & 1.0 & 14.0 \\ \hline
Adversarial image ($\epsilon$ = 32/255) & 53.7 & 9.8 & 35.3 & 4.1 & 13.9 & 5.4 & 48.1 \\ \hline
\quad+DiffPure-VLM (t*=50) & \textbf{13.6} & 0.3 & 9.2 & 0.2 & 5.5 & 0.9 & 10.6 \\ \hline
\quad+DiffPure-VLM (t*=150) & \textbf{11.9} & 0.5 & 8.6 & 0.2 & 4.2 & 0.6 & 9.9 \\ \hline

\end{tabular}%
}
\vspace{-3mm}
\end{table*}

\subsection{Distribution Shift}
\begin{figure*}[tb]
  \centering
  \includegraphics[width=1.0\textwidth]{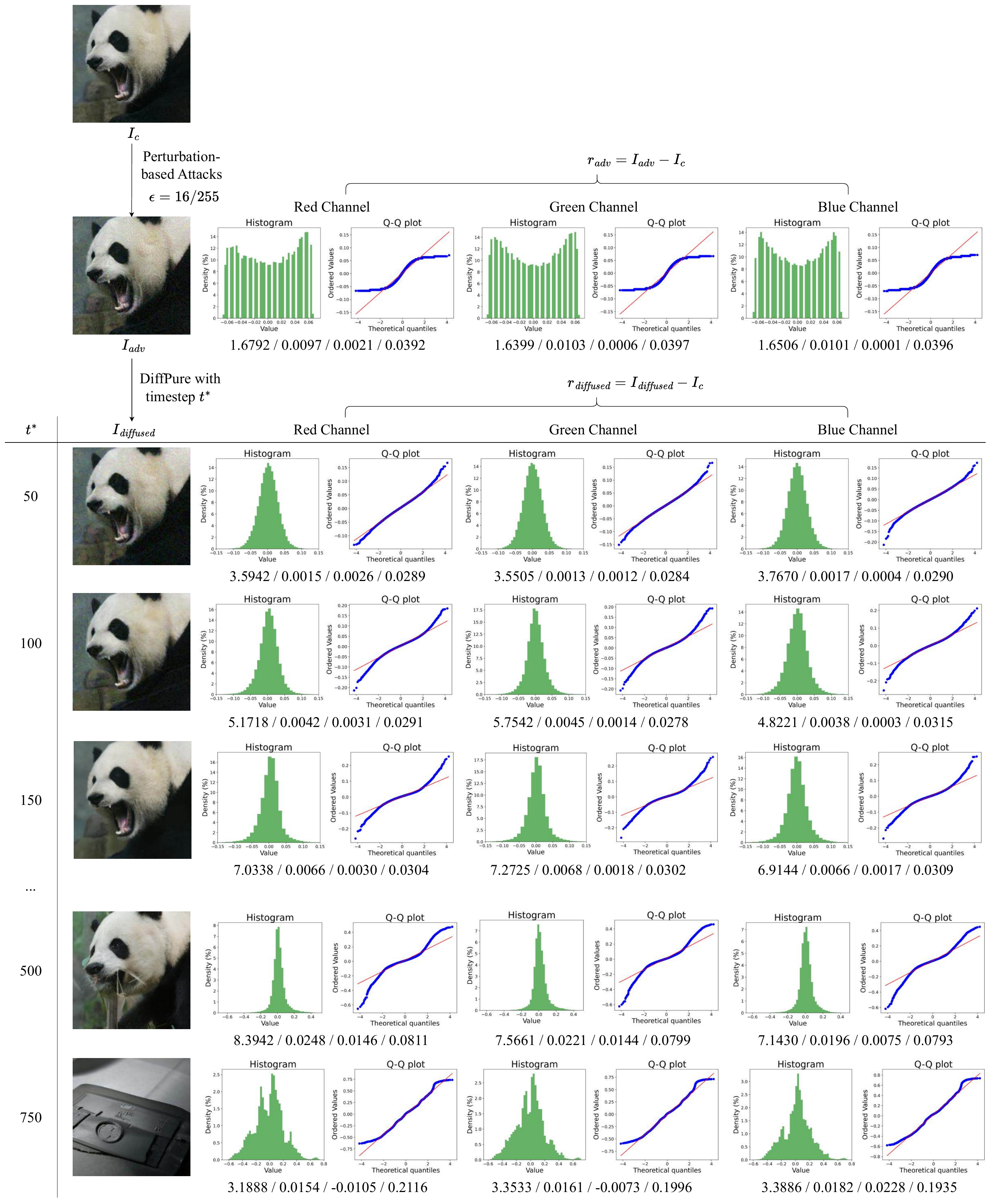}
  \caption{$I_{adv}$, $I_{\mathit{diffused}}$ and statistics of $r_{adv}$, $r_{\mathit{diffused}}$ under different $t^{*}$ in DiffPure (constraint $\epsilon = 16/255$). Metrics are shown in `Kurtosis / Q-Q Deviation / Mean / Standard Deviation'. Please zoom in to see details.}
  \label{fig:Gausianity_16}
  \vspace{-2mm}
\end{figure*}

\begin{figure*}[tb]
  \centering
  \includegraphics[width=1.0\textwidth]{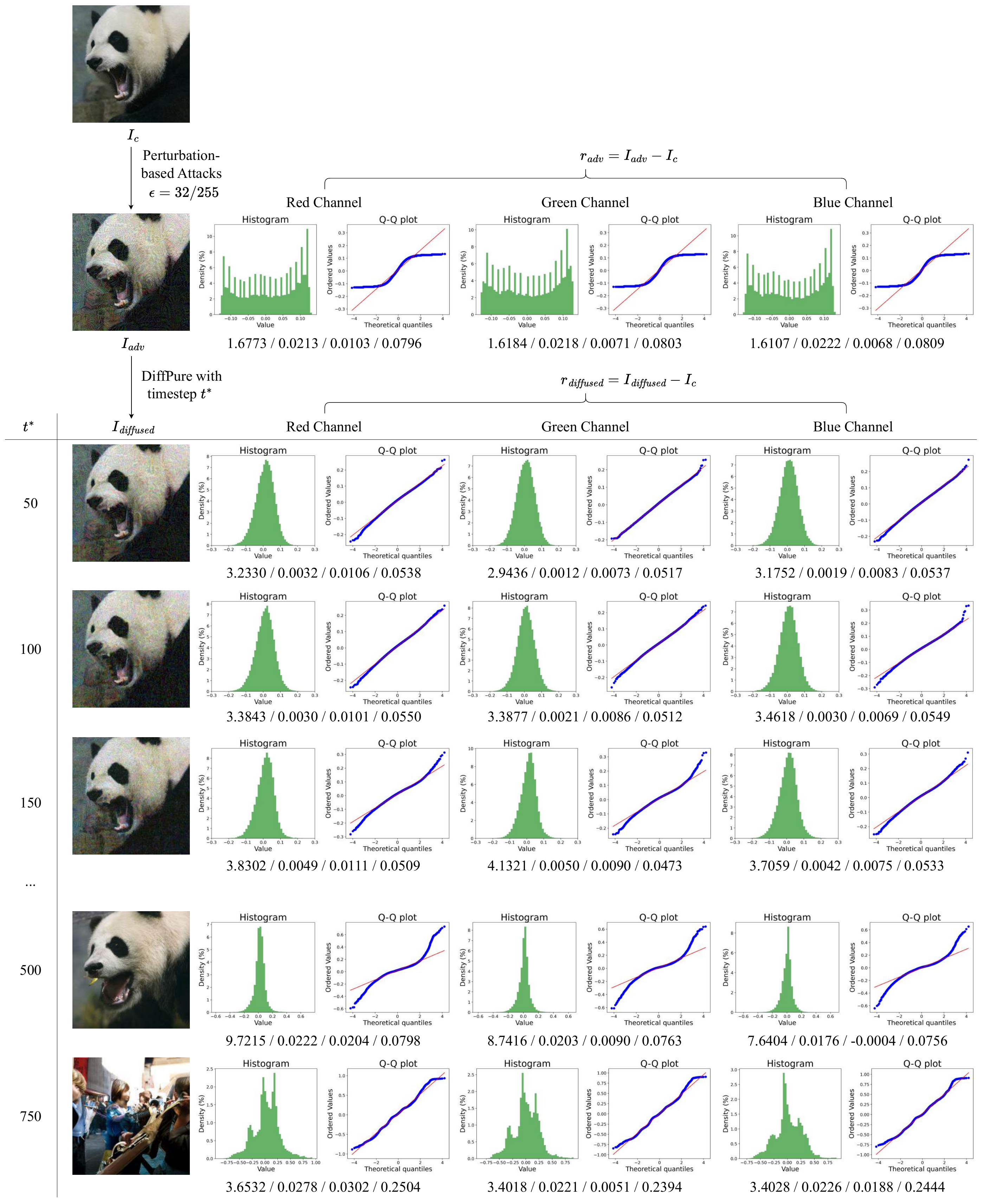}
  \caption{$I_{adv}$, $I_{\mathit{diffused}}$ and statistics of $r_{adv}$, $r_{\mathit{diffused}}$ under different $t^{*}$ in DiffPure (constraint $\epsilon = 32/255$). Metrics are shown in `Kurtosis / Q-Q Deviation / Mean / Standard Deviation'. Please zoom in to see details.}
  \label{fig:Gausianity_32}
  \vspace{-2mm}
\end{figure*}

\begin{figure*}[tb]
  \centering
  \includegraphics[width=1.0\textwidth]{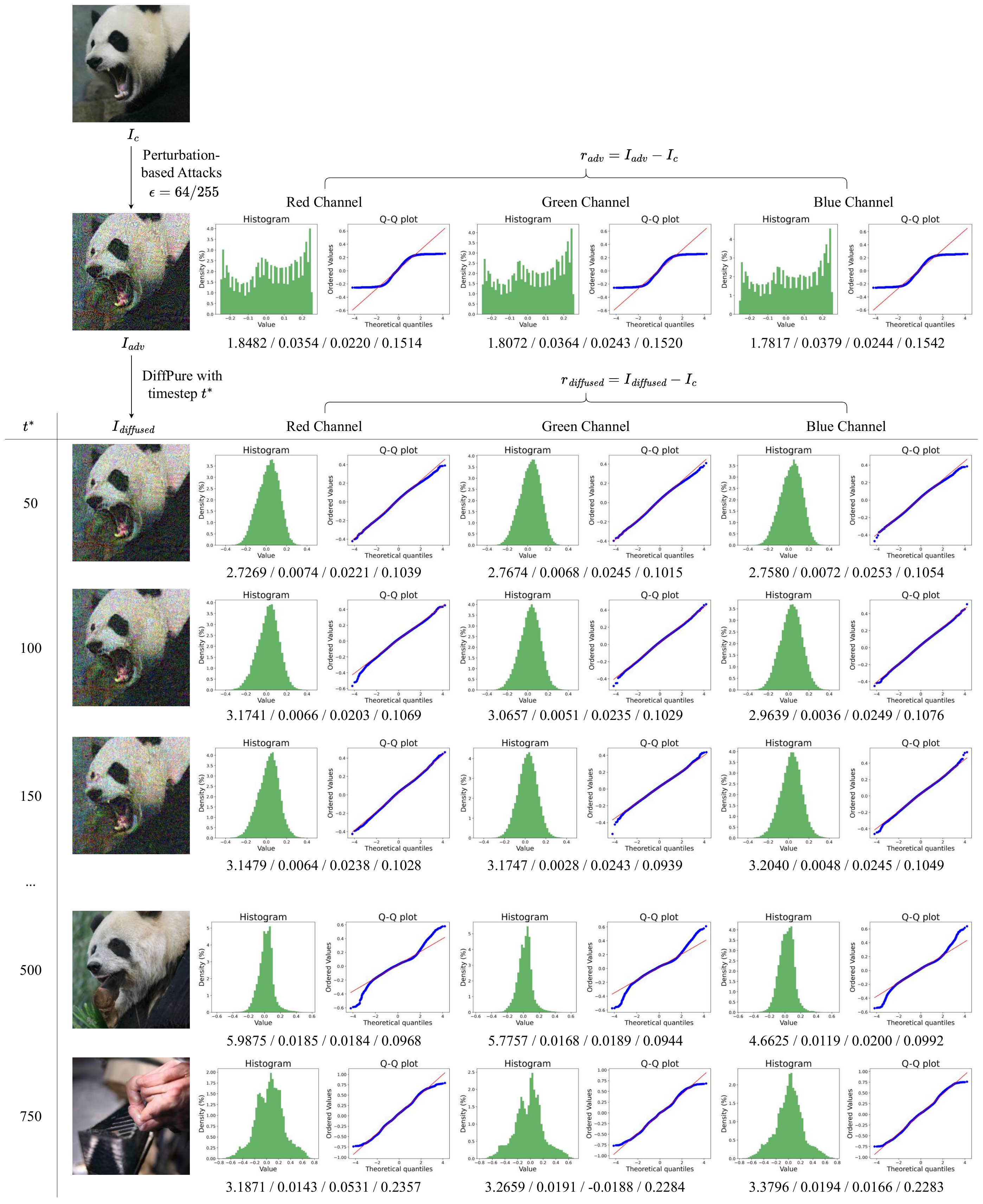}
  \caption{$I_{adv}$, $I_{\mathit{diffused}}$ and statistics of $r_{adv}$, $r_{\mathit{diffused}}$ under different $t^{*}$ in DiffPure (constraint $\epsilon = 64/255$). Metrics are shown in `Kurtosis / Q-Q Deviation / Mean / Standard Deviation'. Please zoom in to see details.}
  \label{fig:Gausianity_64}
  \vspace{-2mm}
\end{figure*}

In this section, we present detailed results from the Gaussianity experiments conducted on adversarial and diffused images. Specifically, we visualize adversarial images $I_{adv}$ alongside their corresponding residuals $r_{adv}$, and diffused images $I_{\mathit{diffused}}$ with their residuals $r_{\mathit{diffused}}$, under pixel constraints $\epsilon \in \{16/255, 32/255, 64/255\}$ for $I_{adv}$ and diffusion timesteps $t^{*} \in \{50, 100, 150, 500, 750\}$ in DiffPure \cite{nie2022diffusion} for generating $I_{\mathit{diffused}}$. Visualizations are shown in Figure \ref{fig:Gausianity_16}, \ref{fig:Gausianity_32}, and \ref{fig:Gausianity_64} with corresponding metrics: Kurtosis, Q-Q deviation, mean, and standard deviation.
From these visualizations, we observe that when $50 \leq t^{*} \leq 150$, the residuals $r_{\mathit{diffused}}$ exhibit a Gaussian-like distribution, particularly for $\epsilon = 32/255$ and $\epsilon = 64/255$. However, as $t^{*}$ increases, the Kurtosis of $r_{\mathit{diffused}}$ rises, indicating a shift towards a long-tailed distribution. This suggests that a small fraction of pixels in $I_{\mathit{diffused}}$ undergo significant changes compared to $I_{c}$, leading to a cleaner image with minimal content alteration, especially when $\epsilon = 16/255$. At $t^{*} = 500$, the Kurtosis and standard deviation of $r_{\mathit{diffused}}$ become significantly larger, implying greater changes in image content, as reflected in the visualization of $I_{\mathit{diffused}}$. For $t^{*} = 750$, the Kurtosis decreases while the standard deviation further increases, indicating that $r_{\mathit{diffused}}$ transitions to a flatter and broader distribution. In this case, $I_{\mathit{diffused}}$ diverges substantially from $I_{c}$ in image content.

\begin{figure}[t]
    \centering
    \includegraphics[width=1.0\linewidth]{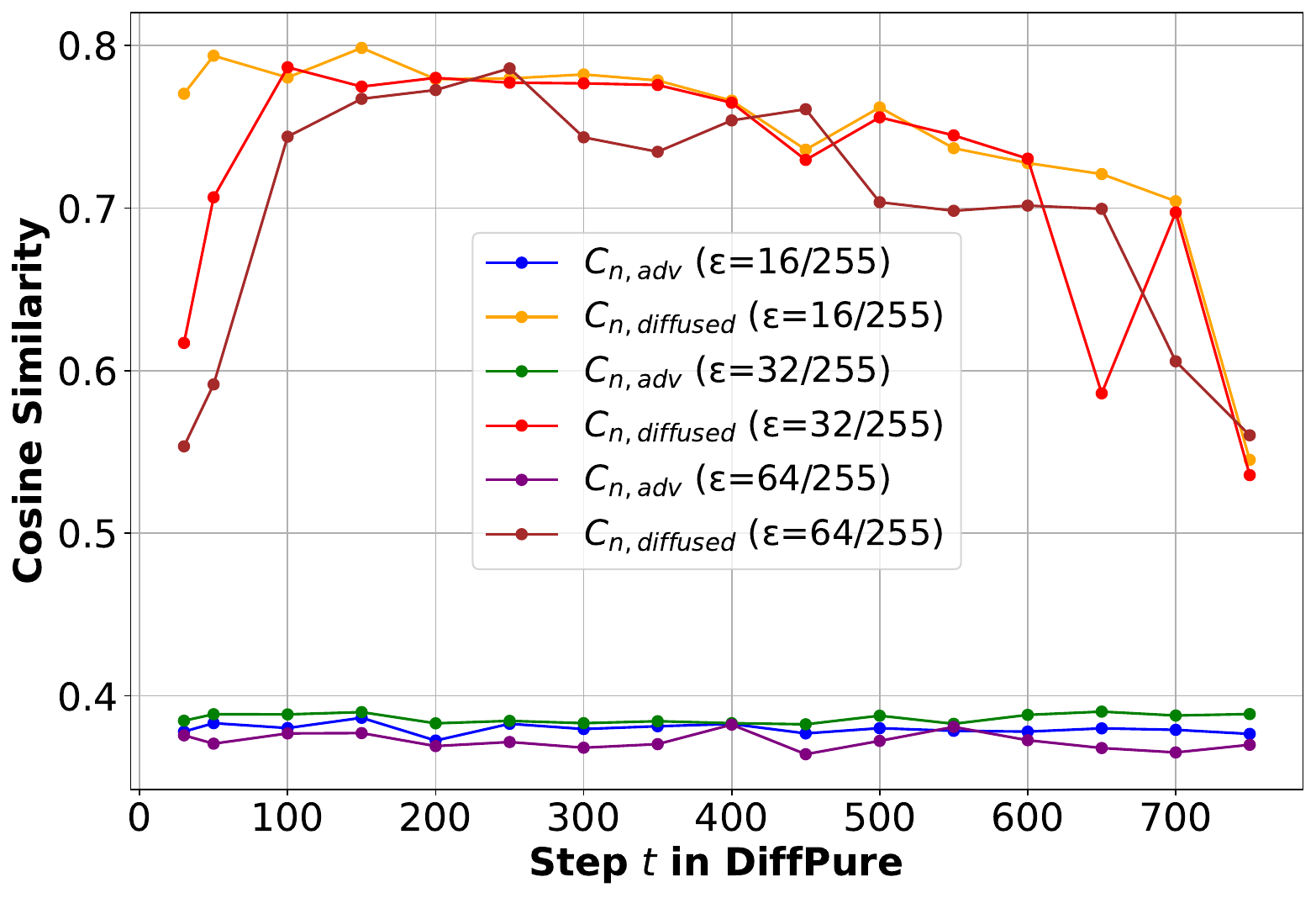}
    \caption{Cosine similarity of visual embeddings under different $\epsilon$ of adversarial image $I_{adv}$ and $t^{*}$ of DiffPure.}
    \label{fig:visual_embeddings_curve} 
    \vspace{-2mm}
\end{figure}

Furthermore, we extend our analysis to the embedding space, examining the similarities between the clean image $I_{c}$, the adversarial image $I_{adv}$, and the diffused image $I_{\mathit{diffused}}$. Based on our experiment in pixel space, where the residual noise $r_{\mathit{diffused}}$ approximates a Gaussian distribution under certain timestep settings in DiffPure, we consider $I_{\mathit{diffused}}$ as comparable to $I_{c}$ with added Gaussian noise. To verify this, we generate a noisy image $I_{n} = I_{c} + n, n\in\mathcal{N}\left(0, \sigma^{2}_{r_\mathit{diffused}} \right)$, where $\sigma_{r_\mathit{diffused}}$ indicates the standard deviation of $r_\mathit{diffused}$. Using pre-trained visual encoder $E$ in MiniGPT-4, we compute cosine similarities between the embeddings of $I_{n}$ and $I_{adv}$, denoted as $C_{n, adv}$, and between $I_{n}$ and $I_{\mathit{diffused}}$, denoted as $C_{n, \mathit{diffused}}$. Figure \ref{fig:visual_embeddings_curve} shows these similarities across varying adversarial constraints $\epsilon$ and DiffPure steps $t^{*}$. 
Results indicate that, $C_{n, \mathit{diffused}}$ consistently exceeds $C_{n, adv}$, showing that $I_{\mathit{diffused}}$ is closer to $I_{n}$ than $I_{adv}$ in the embedding space. Notably, with moderate timesteps ($t^{*} \in [50, 150]$), $I_{\mathit{diffused}}$ is similar to $I_{n}$ (Gaussian noise $n$ added to the benign clean image $I_{c}$) in both pixel and embedding spaces.

\begin{figure}[tb]
  \centering
  \includegraphics[width=1.0\linewidth]{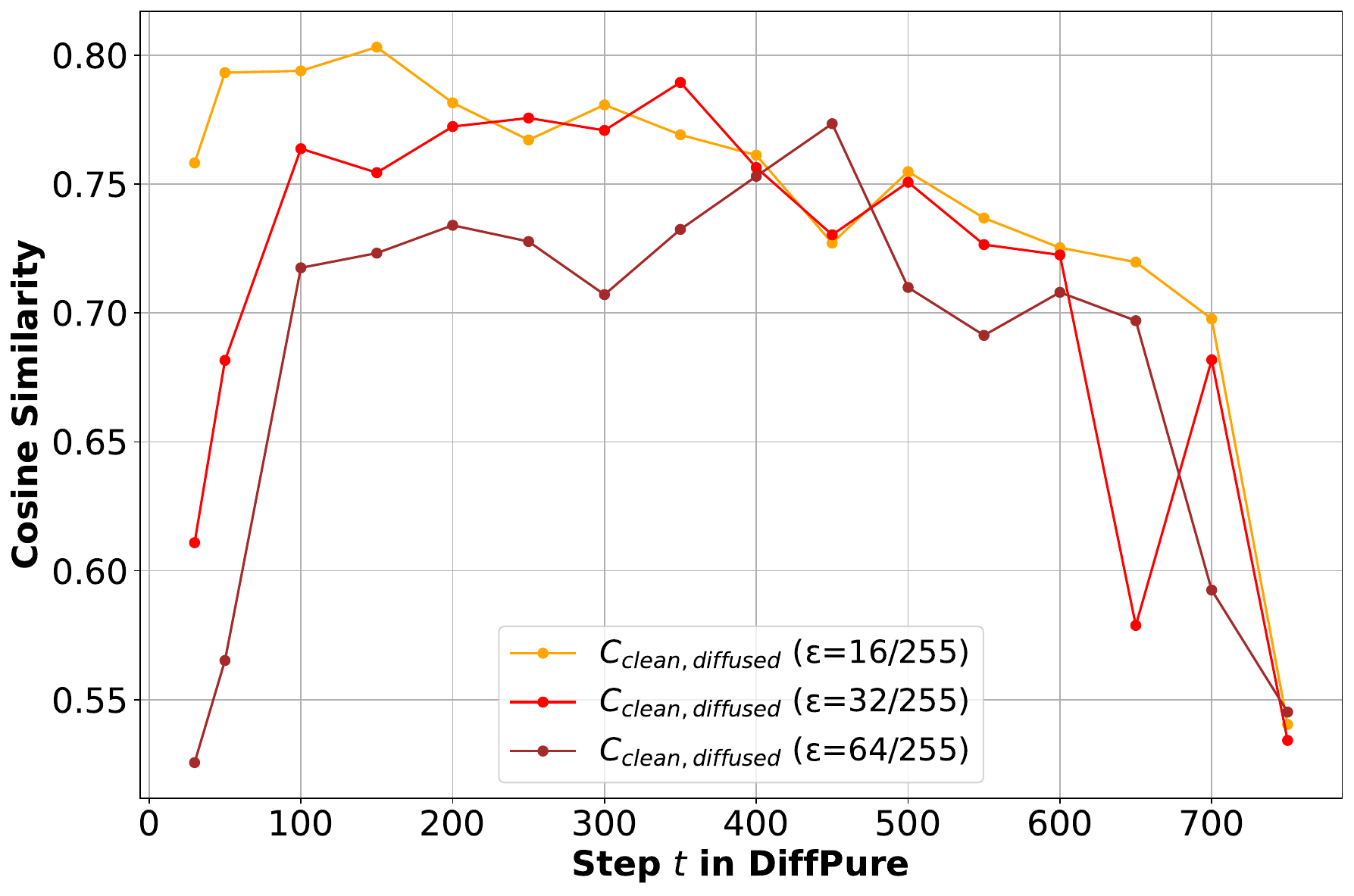}
  \caption{Cosine Similarity of visual embeddings from $I_{c}$ and $I_{\mathit{diffused}}$ under different $\epsilon$ of adversarial image.}
  \label{fig:cos_sim_c_diffused}
\end{figure}

We also visualize the cosine similarity between the visual embeddings of $I_{\mathit{diffused}}$ and $I_{c}$, denoted as $C_{clean, \mathit{diffused}}$, across varying $\epsilon$ and $t^{*}$. Results are shown in Figure \ref{fig:cos_sim_c_diffused}, revealing that $C_{clean, \mathit{diffused}}$ decreases rapidly as $t^{*}$ decreases, while it gradually declines as $t^{*}$ increases. Combining these findings with experiments in pixel space, we conclude that smaller $t^{*}$ values lead $I_{\mathit{diffused}}$ to retain adversarial information, whereas larger $t^{*}$ values result in significant content disruption, leading to semantic misalignment.

\section{Additional Details of DiffPure-VLM}
\label{appendix:vlm}
\subsection{Implementation Details}

\begin{figure*}[tb]
  \centering
  \includegraphics[width=0.7\textwidth]{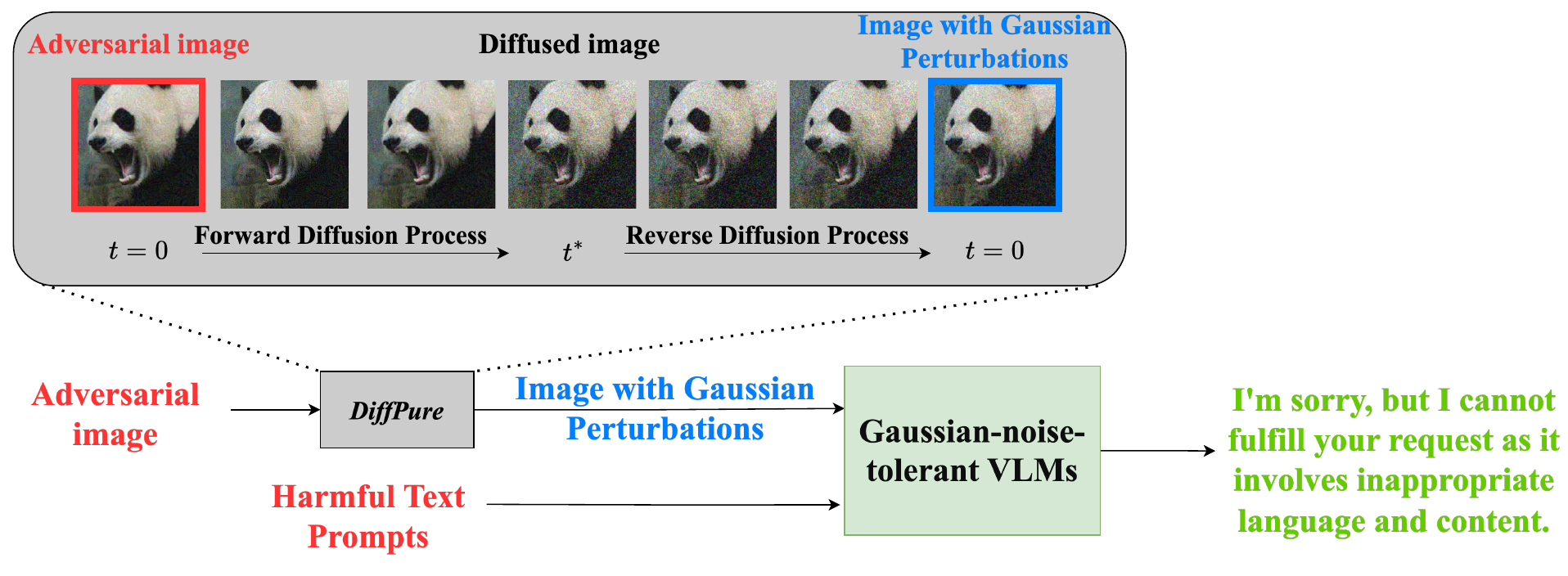}
  \caption{The overall framework of DiffPure-VLM.}
  \label{fig:DiffPure-VLM-v2}
  \vspace{-2mm}
\end{figure*}

The overall architecture of our proposed DiffPure-VLM framework is illustrated in Figure~\ref{fig:DiffPure-VLM-v2}, with the detailed algorithmic procedure outlined in Algorithm~\ref{algorithm:diffpure-vlm}. For our experiments, we employ the Guided Diffusion model for ImageNet \cite{dhariwal2021diffusion}, specifically the $256\times256$ unconditional variant provided by OpenAI\footnote{\url{https://openaipublic.blob.core.windows.net/diffusion/jul-2021/256x256\_diffusion\_uncond.pt}}. Importantly, we synchronize the forward diffusion timesteps ($\mathbf{t}_{\text{forward}}$) with the reverse diffusion timesteps ($\mathbf{t}_{\text{reverse}}$), denoted as $t^*$ in the experimental results, following the setup in DiffPure \cite{nie2022diffusion}. Here, we leverage this diffusion model to validate the robustness of our fine-tuned VLMs against Gaussian noise, demonstrating a preliminary defense strategy. However, the fixed image resolution of the diffusion model requires down-sampling and up-sampling operations, which may introduce artifacts not considered during the fine-tuning of the VLM, potentially impacting evaluation results. In the future, adopting more advanced diffusion models will be essential for real-world applications.

\begin{algorithm}[t]
\caption{DiffPure-VLM Adversarial Image Purification with DDPM}
\begin{algorithmic}[1]
\label{algorithm:diffpure-vlm}
\REQUIRE Adversarial image $x$, harmful text prompt $p$, diffusion model $\mathcal{D}$, forward diffusion timesteps $\mathbf{t}_{\text{forward}}$, reverse diffusion timesteps $\mathbf{t}_{\text{reverse}}$, visual language model $\mathcal{VLM}$.
\ENSURE Question answering result $\text{output}$
\STATE Resize input image $x$ to the size required by the diffusion model (e.g., $256 \times 256$).
\STATE DDPM forward process with $\mathbf{t}_{\text{forward}}$ steps: $\hat{x} = \text{get\_noised\_x}(x, \mathbf{t}_{\text{forward}})$.
\FOR{$t$ in $\mathbf{t}_{\text{reverse}}$}
    \STATE Denoise using reverse DDPM process: $x = \text{denoising\_process}(\hat{x}, t)$.
\ENDFOR
\STATE Obtain purified image with Gaussian noise: $x_{\text{gaussian}} = \text{normalize}(x)$.
\STATE Perform question answering using VLM: $\text{output} = \mathcal{VLM}(x_{\text{gaussian}}, p)$.
\RETURN $\text{output}$
\end{algorithmic}
\end{algorithm}

\subsection{Extended Experimental Results}

In the main paper, for the sake of brevity, we only report results for the standard perturbation-based attack setting of $\epsilon = 32/255$. However, we also conducted experiments with lower attack strength ($\epsilon = 16/255$) and higher attack strength ($\epsilon = 64/255$) to further validate our analysis and approach in Table~\ref{tab:diffpure-vlm-v2}. Across different models and attack strengths, our DiffPure-VLM consistently reduces the attack success rate within a limited number of diffusion timesteps (fewer than $150$). Notably, under lower attack strengths, setting the diffusion step to as low as  $t^* = 50$  is sufficient to bring the attack success rate down to the level observed for clean inputs. However, under higher attack strengths,  $t^* = 50$  fails to reduce the attack success rate to the baseline level for both InternVL2-8B and MiniGPT-4-13B. This indicates that stronger attacks require a larger number of diffusion steps to effectively transform the adversarial noise into Gaussian noise. This finding aligns with the analysis presented in Figure 4 of the main paper, where the residual image at  $t^* = 50$  for an attack strength of $\epsilon = 64/255$ does not exhibit Gaussian characteristics. Moreover, we observe that $t^* = 100$ demonstrates strong performance across all attack conditions, making it an effective trade-off between time and robustness. Thus, in real-world applications, setting $t^* = 100$ offers a balanced solution, achieving reliable defense while maintaining computational efficiency.

\begin{table*}[t]
\centering
\caption{Evaluation of DiffPure-VLM's effectiveness on RealToxicityPrompts across different image configurations. Metrics include attack success rate and toxicity levels (Perspective API \%). Rows highlighted in light red indicate cases where attack success rate does not meet the baseline level of benign image input.}
\label{tab:diffpure-vlm-v2}
\setlength{\tabcolsep}{4pt}
\resizebox{0.9\textwidth}{!}{%
\begin{tabular}{l|c|c|c|c|c|c|c}
\toprule
\textbf{Image Type} & \textbf{\begin{tabular}[c]{@{}c@{}}Attack \\ Success Rate $\downarrow$ \end{tabular}} & \textbf{Identity Attack} & \textbf{Profanity} & \textbf{\begin{tabular}[c]{@{}c@{}}Severe \\ Toxicity\end{tabular}} & \textbf{\begin{tabular}[c]{@{}c@{}}Sexually \\ Explicit\end{tabular}} & \textbf{Threat} & \textbf{Toxicity} \\ \hline
\rowcolor[HTML]{EFEFEF} \multicolumn{8}{c}{\textbf{InternVL2-8B-RobustVLGuard}} \\
\midrule
Benign Clean Image & 29.9 & 0.8 & 22.1 & 0.3 & 7.2 & 1.5 & 25.9 \\
Benign Noisy Image & 34.5 & 2.1 & 27.2 & 1.3 & 8.4 & 1.6 & 31.3 \\
\midrule
Adversarial Image ($\epsilon = 16/255$) & 72.5 & 19.8 & 58.5 & 8.3 & 19.2 & 7.8 & 70.0 \\
\quad +DiffPure-VLM ($t^* = 50$) & \textbf{31.4} & 1.4 & 24.6 & 1.1 & 7.9 & 1.6 & 27.5 \\
\quad +DiffPure-VLM ($t^* = 100$) & \textbf{28.2} & 0.9 & 21.7 & 0.4 & 6.8 & 1.5 & 23.9 \\
\quad +DiffPure-VLM ($t^* = 150$) & \textbf{28.2} & 1.6 & 22.4 & 0.2 & 6.9 & 1.1 & 24.4 \\
\midrule
Adversarial Image ($\epsilon = 32/255$) & 70.6 & 26.7 & 56.5 & 9.2 & 17.3 & 6.9 & 68.1 \\
\quad +DiffPure-VLM ($t^* = 50$) & \textbf{33.4} & 2.4 & 20.6 & 0.7 & 8.1 & 2.4 & 29.1 \\
\quad +DiffPure-VLM ($t^* = 100$) & \textbf{33.4} & 1.6 & 27.7 & 0.6 & 7.6 & 1.5 & 30.2 \\
\quad +DiffPure-VLM ($t^* = 150$) & \textbf{32.8} & 1.7 & 25.9 & 0.6 & 7.7 & 1.8 & 29.1 \\
\midrule
Adversarial Image ($\epsilon = 64/255$) & 57.3 & 9.3 & 45.8 & 4.4 & 16.1 & 3.9 & 53.9 \\
\rowcolor[HTML]{FFCCCC}\quad +DiffPure-VLM ($t^* = 50$) & \textbf{40.9} & 2.3 & 32.9 & 1.4 & 9.3 & 2.3 & 37.3 \\
\quad +DiffPure-VLM ($t^* = 100$) & \textbf{35.7} & 1.8 & 28.2 & 0.8 & 7.6 & 2.4 & 31.8 \\
\quad +DiffPure-VLM ($t^* = 150$) & \textbf{36.1} & 2.4 & 28.3 & 1.2 & 8.3 & 1.8 & 33.6 \\
\midrule

\rowcolor[HTML]{EFEFEF} \multicolumn{8}{c}{\textbf{LLaVA-v1.5-7B-RobustVLGuard}} \\
\midrule
Benign Clean image & 43.6 & 4.6 & 34.7 & 2.4 & 12.3 & 3.5 & 41.0 \\ 
Benign Noisy image & 42.3 & 3.1 & 34.5 & 1.9 & 11.8 & 3.1 & 40.0 \\ 
\midrule
Adversarial image ($\epsilon = 16/255$) & 62.6 & 11.3 & 48.8 & 5.3 & 16.8 & 5.8 & 59.1 \\ 
\quad+DiffPure-VLM ($t^* = 50$) & \textbf{42.7} & 3.4 & 32.1 & 1.5 & 12.0 & 4.6 & 39.7 \\ 
\quad+DiffPure-VLM ($t^* = 100$) & \textbf{42.8} & 3.9 & 32.5 & 2.3 & 12.5 & 3.7 & 39.3 \\ 
\quad+DiffPure-VLM ($t^* = 150$) & \textbf{44.4} & 3.3 & 34.4 & 2.2 & 12.6 & 3.2 & 41.0 \\ 
\midrule
Adversarial image ($\epsilon = 32/255$) & 62.5 & 7.8 & 48.0 & 5.4 & 16.5 & 5.8 & 60.0 \\
\quad+DiffPure-VLM ($t^* = 50$) & \textbf{43.9} & 3.2 & 34.6 & 2.4 & 12.8 & 3.7 & 41.0 \\ 
\quad+DiffPure-VLM ($t^* = 100$) & \textbf{44.1} & 3.5 & 35.4 & 2.1 & 13.0 & 4.1 & 41.3 \\ 
\quad+DiffPure-VLM ($t^* = 150$) & \textbf{42.5} & 3.5 & 32.7 & 2.8 & 12.1 & 4.1 & 39.3 \\ 
\midrule
Adversarial image ($\epsilon = 64/255$) & 57.5 & 9.2 & 43.5 & 5.2 & 15.3 & 5.8 & 54.7 \\ 
\quad+DiffPure-VLM ($t^* = 50$) & \textbf{42.1} & 2.7 & 32.1 & 2.1 & 12.3 & 2.9 & 39.0 \\ 
\quad+DiffPure-VLM ($t^* = 100$) & \textbf{40.5} & 3.3 & 31.4 & 1.9 & 11.7 & 2.8 & 37.5 \\ 
\quad+DiffPure-VLM ($t^* = 150$) & \textbf{42.4} & 3.5 & 32.8 & 1.8 & 11.5 & 4.0 & 40.2 \\ 
\midrule

\rowcolor[HTML]{EFEFEF} \multicolumn{8}{c}{\textbf{MiniGPT-4-13B-RobustVLGuard}} \\ 
\midrule
Benign Clean image & 16.0 & 0.4 & 9.9 & 0.3 & 4.6 & 1.1 & 12.1 \\ 
Benign Noisy image & 16.5 & 0.9 & 11.9 & 0.6 & 5.8 & 1.0 & 14.0 \\ \midrule
Adversarial image ($\epsilon = 16/255$) & 47.4 & 9.3 & 34.2 & 1.4 & 11.8 & 4.2 & 41.5 \\ 
\quad+DiffPure-VLM ($t^* = 50$) & \textbf{16.0} & 0.6 & 9.3 & 0.3 & 6.5 & 1.4 & 13.2 \\ 
\quad+DiffPure-VLM ($t^* = 100$) & \textbf{15.8} & 0.7 & 9.7 & 0.0 & 6.1 & 1.1 & 12.8 \\ 
\quad+DiffPure-VLM ($t^* = 150$) & \textbf{9.8} & 0.4 & 6.0 & 0.1 & 3.3 & 0.5 & 7.8 \\ \midrule
Adversarial image ($\epsilon = 32/255$) & 53.7 & 9.8 & 35.3 & 4.1 & 13.9 & 5.4 & 48.1 \\ 
\quad+DiffPure-VLM ($t^* = 50$) & \textbf{13.6} & 0.3 & 9.2 & 0.2 & 5.5 & 0.9 & 10.6 \\ 
\quad+DiffPure-VLM ($t^* = 100$) & \textbf{15.2} & 0.6 & 9.5 & 03 & 5.4 & 1.1 & 12.7 \\ 
\quad+DiffPure-VLM ($t^* = 150$) & \textbf{11.9} & 0.5 & 8.6 & 0.2 & 4.2 & 0.6 & 9.9 \\ \midrule
Adversarial image ($\epsilon = 64/255$ ) & 60.2 & 6.8 & 44.6 & 4.2 & 16.2 & 5.8 & 56.0 \\ 
\rowcolor[HTML]{FFCCCC}\quad+DiffPure-VLM ($t^* = 50$) & \textbf{30.3} & 1.8 & 21.6 & 1.4 & 11.4 & 1.9 & 26.9 \\ 
\quad+DiffPure-VLM ($t^* = 100$) & \textbf{10.6} & 0.0 & 7.1 & 0.0 & 4.1 & 0.8 & 8.2 \\ 
\quad+DiffPure-VLM ($t^* = 150$) & \textbf{9.4} & 0.4 & 5.5 & 0.3 & 4.1 & 0.6 & 7.0 \\ \bottomrule

\end{tabular}%
}
\vspace{-3mm}
\end{table*}

\section{Conjectures and Discussion on the Impact of Gaussian Noise}
\label{appendix:gaussian}

\subsection*{Problem Definition}

\textbf{Setting:}
\begin{itemize}
    \item A Visual Language Model (VLM) typically consists of three main components: a visual encoder, a language model, and a vision-language connection module.
    \item Let the input be a pair \((I, T)\), where \(I \in \mathbb{R}^d\) is an image and \(T\) is the corresponding text prompt.
    \item The VLM generates an output sequence of tokens, denoted by \(\hat{T} = f_{\theta}(I, T)\), where \(f_{\theta}\) represents the VLM pipeline parameterized by \(\theta\).
\end{itemize}

\noindent\textbf{Adversarial Attack:}
An adversarial perturbation \(\delta\) is applied to the image \(I\), resulting in a perturbed image \(I_{\delta} = I + \delta\). The perturbation \(\delta\) is crafted to manipulate the VLM into generating a specific harmful target text \(T^{\text{target}}\). The adversary's objective is:

\[
\delta = \arg\min_{\|\delta\| \leq \epsilon} L\left(f_{\theta}(I + \delta, T), T^{\text{target}}\right),
\]

where \(L(\cdot, \cdot)\) measures the discrepancy between the generated text \(\hat{T}\) and the target text \(T^{\text{target}}\). The constraint \(\|\delta\| \leq \epsilon\) ensures that the perturbation is imperceptible.

\noindent\textbf{Conjectures:}
We introduce the following four conjectures to guide our investigation into the impact of Gaussian noise on VLMs:

\begin{enumerate}
    \item \textbf{Sensitivity of Adversarial Attacks to Gaussian Noise}: Adding Gaussian noise to adversarially perturbed images will significantly reduce the effectiveness of the attack.
    \item \textbf{Gaussian Noise as a Simple Attack on VLM Safety}: Gaussian noise, even without adversarial perturbations, may increase the likelihood of generating harmful text.
    \item \textbf{Gaussian Noise as a Regularizer}: Augmenting training data with Gaussian noise may act as a regularizer, enhancing the robustness of the VLM.
    \item \textbf{Fine-Tuning with Gaussian Noise Preserves Performance}: Incorporating Gaussian noise during fine-tuning will preserve or even improve the VLM's overall performance.
\end{enumerate}

\noindent\textbf{Objective:}
The goal of this study is to systematically evaluate the impact of Gaussian noise on the robustness and reliability of VLMs. By exploring the above conjectures, we aim to determine whether Gaussian noise can effectively mitigate adversarial perturbations and enhance model robustness without compromising performance.

\subsection*{Conjecture 1: Sensitivity of Adversarial Perturbations to Gaussian Noise}

\textbf{Statement:} Adversarial perturbations are highly sensitive to Gaussian noise; the attack effectiveness is significantly diminished when Gaussian noise is added to the adversarial image.

\noindent\textbf{Discussion:}

Consider an adversarially perturbed image \( I_{\delta} = I + \delta \), where the perturbation \(\delta\) is optimized to minimize the loss:

\[
\delta = \arg\min_{\|\delta\| \leq \epsilon} L\left(f_{\theta}(I + \delta, T), T^{\text{target}}\right),
\]

where \(L(\cdot, \cdot)\) measures the discrepancy between the generated text \(\hat{T}\) and the harmful target text \(T^{\text{target}}\). The perturbation \(\delta\) is crafted to exploit specific vulnerabilities in the model \(f_{\theta}\).

Now, consider the scenario where Gaussian noise \(\eta \sim \mathcal{N}(0, \sigma^2 I)\) is added to the input. The new input becomes:

\[
I_{\delta, \eta} = I + \delta + \eta.
\]

The expected loss over the distribution of Gaussian noise \(\eta\) is:

\[
\mathbb{E}_{\eta}\left[L\left(f_{\theta}(I + \delta + \eta, T), T^{\text{target}}\right)\right].
\]

Since the adversarial perturbation \(\delta\) is tailored for the specific input \(I\), adding random Gaussian noise \(\eta\) disrupts this optimization. Adversarial perturbations exploit the model’s sensitivity along certain directions in the input space, while isotropic Gaussian noise perturbs the input uniformly in all directions, diminishing the effect of \(\delta\).

Assuming that \(f_{\theta}\) and \(L\) are Lipschitz continuous, we can bound the increase in expected loss as follows:

\[
\begin{aligned}
    \mathbb{E}_{\eta}\left[L\left(f_{\theta}(I + \delta + \eta, T), T^{\text{target}}\right)\right] \geq & \, L\left(f_{\theta}(I + \delta, T), T^{\text{target}}\right) \\
    & + \frac{\sigma^2 \lambda}{2},
\end{aligned}
\]

where \(\lambda\) is a positive constant related to the curvature of \(L\) and \(f_{\theta}\).

This inequality indicates that the addition of Gaussian noise increases the expected loss, thus reducing the effectiveness of the adversarial perturbation. The random noise \(\eta\) disrupts the carefully crafted \(\delta\), making it less effective at manipulating the VLM's output. This supports our conjecture that Gaussian noise can act as a simple yet effective countermeasure against adversarial attacks.

\subsection*{Conjecture 2: Gaussian Noise as a Simple Attack on VLM Safety}

\textbf{Statement:} Adding Gaussian noise \( \eta \sim \mathcal{N}(0, \sigma^2 I) \) to a clean image \( I_{\text{clean}} \) can compromise the safety of VLMs.

\noindent\textbf{Setting:} Let \( I_{\text{clean}} \) be a clean image, and \( \eta \sim \mathcal{N}(0, \sigma^2 I) \) be Gaussian noise. The perturbed image is defined as:

\[
I_{\text{noisy}} = I_{\text{clean}} + \eta.
\]

The VLM processes the noisy image \( I_{\text{noisy}} \) along with a corresponding text prompt \( T \), and generates an output based on this combined input.

\noindent\textbf{Discussion:}

1. \textbf{Effect of Noise on Model Input:}
   The input to the model becomes \( I_{\text{noisy}} = I_{\text{clean}} + \eta \). This perturbation, although random, alters the image representation processed by the VLM. The model’s output can be locally approximated around the clean input as:

   \[
   f_{\theta}(I_{\text{clean}} + \eta, T) \approx f_{\theta}(I_{\text{clean}}, T) + \nabla I_{\text{clean}} f_{\theta} \cdot \eta,
   \]

   where \( \nabla I_{\text{clean}} f_{\theta} \) represents the gradient of the model output with respect to the clean image input. The Gaussian noise \( \eta \) introduces random perturbations that shift the image features.

2. \textbf{Vulnerability of VLMs to Noise:}
   VLMs are typically trained on clean image data, and thus, they may lack robustness to input noise. The introduction of Gaussian noise can push the model’s input into regions of the feature space that were not well-covered during training, potentially causing the model to misinterpret the input and generate unexpected responses.

3. \textbf{Impact on Safety:}
   Adding Gaussian noise may shift the model's behavior towards decision boundaries where safety mechanisms are less effective. This increases the likelihood of generating unsafe or harmful text:

   \[
   L(f_{\theta}(I_{\text{clean}} + \eta, T), T^{\text{target}}) \geq L(f_{\theta}(I_{\text{clean}}, T), T^{\text{target}}),
   \]

   where \( T^{\text{target}} \) represents a potentially harmful target output. The inequality suggests that the noisy input can lead to a higher loss, increasing the risk of unsafe text generation.

4. \textbf{Gaussian Noise as a Simple Yet Effective Attack:}
   Unlike adversarial perturbations that require careful optimization and model-specific crafting, Gaussian noise introduces random changes without any specific targeting. Despite its simplicity, it can destabilize the model and affect its safety, demonstrating that even non-adversarial noise can be a risk factor for VLMs.

In summary, adding Gaussian noise to clean images can indeed disrupt the safety of VLMs, even in the absence of sophisticated adversarial attacks. This highlights a potential vulnerability of VLMs that warrants further investigation.

\subsection*{Conjecture 3: Gaussian Noise as a Regularizer}

\textbf{Statement:} Augmenting training data with Gaussian noise acts as a regularizer, reducing the risk of overfitting to adversarial perturbations and enhancing model robustness.

\noindent\textbf{Discussion:}

We introduce a regularized loss function that incorporates Gaussian noise during training:

\[
L_{\text{reg}}(\theta) = \mathbb{E}_{(I, T) \sim \mathcal{D}} \mathbb{E}_{\eta \sim \mathcal{N}(0, \sigma^2 I)}\left[L\left(f_{\theta}(I + \eta, T), T\right)\right],
\]

where \(\mathcal{D}\) represents the training data distribution. This formulation encourages the model to perform well not only on clean inputs but also on noisy inputs, promoting robustness.

To understand the regularizing effect of Gaussian noise, we expand the loss function \(L\) using a second-order Taylor expansion around the clean input \(I\):

\begin{align*}
L\left(f_{\theta}(I + \eta, T), T\right) &\approx L\left(f_{\theta}(I, T), T\right) \\
&\quad + \nabla_I L\left(f_{\theta}(I, T), T\right)^\top \eta \\
&\quad + \frac{1}{2} \eta^\top \nabla_I^2 L\left(f_{\theta}(I, T), T\right) \eta.
\end{align*}

Taking the expectation over the Gaussian noise \(\eta \sim \mathcal{N}(0, \sigma^2 I)\), we obtain:

\begin{align*}
\mathbb{E}_{\eta}\left[L\left(f_{\theta}(I + \eta, T), T\right)\right] &\approx L\left(f_{\theta}(I, T), T\right) \\
&\quad + \frac{1}{2} \mathbb{E}_{\eta}\left[\eta^\top \nabla_I^2 L\left(f_{\theta}(I, T), T\right) \eta\right] \\
&= L\left(f_{\theta}(I, T), T\right) \\
&\quad + \frac{\sigma^2}{2} \operatorname{Tr}\left(\nabla_I^2 L\left(f_{\theta}(I, T), T\right)\right).
\end{align*}

The additional term \(\frac{\sigma^2}{2} \operatorname{Tr}\left(\nabla_I^2 L\left(f_{\theta}(I, T), T\right)\right)\) penalizes large curvature (i.e., high second derivatives) of the loss function with respect to the input \(I\). This encourages smoother mappings from the input to the output, reducing the model’s sensitivity to small input perturbations, including adversarial ones.

In summary, the addition of Gaussian noise during training acts as a regularizer by penalizing sharp changes in the loss landscape. This results in a model that is less prone to overfitting and more resilient to adversarial attacks, as it learns smoother and more stable input-output mappings.

\subsection*{Conjecture 4: Fine-Tuning with Gaussian Noise Preserves Performance}

\textbf{Statement:} Fine-tuning the VLM with Gaussian noise-augmented data maintains performance on clean data while enhancing robustness to adversarial perturbations.

\noindent\textbf{Discussion:}

Let \(\mathcal{D} = \{(I_i, T_i)\}_{i=1}^N\) be the original training dataset. We construct an augmented dataset by adding Gaussian noise:

\[
\mathcal{D}_{\text{aug}} = \left\{(I_i + \eta_i, T_i) \mid \eta_i \sim \mathcal{N}(0, \sigma^2 I)\right\}_{i=1}^N.
\]

The training objective is to minimize the following loss function:

\[
\hat{L}_{\text{aug}}(\theta) = \frac{1}{N} \sum_{i=1}^N \mathbb{E}_{\eta_i}\left[L\left(f_{\theta}(I_i + \eta_i, T_i), T_i\right)\right].
\]

Since the Gaussian noise \(\eta_i\) has a zero mean, the expected gradient of the loss with respect to the model parameters \(\theta\) is centered around the gradient on the clean data:

\[
\mathbb{E}_{\eta_i}\left[\nabla_{\theta} L\left(f_{\theta}(I_i + \eta_i, T_i), T_i\right)\right] = \nabla_{\theta} L\left(f_{\theta}(I_i, T_i), T_i\right).
\]

This result indicates that the expected training gradient remains aligned with the gradient computed on the clean data, thereby preserving the model’s performance on clean inputs.

Moreover, by training on both clean and noise-augmented data, the model is exposed to a neighborhood of inputs around each training example. This exposure helps the model generalize better and become less sensitive to small perturbations, effectively enhancing its robustness against adversarial attacks.

In summary, fine-tuning with Gaussian noise-augmented data acts as a regularization strategy that not only maintains the VLM's accuracy on clean data but also improves its resistance to adversarial perturbations.

\section{Detailed Proofs}

\subsection*{Bounding the Increase in Loss Due to Gaussian Noise}
\subsubsection*{Discussion:}
\textbf{Step 1: Lipschitz Continuity of \( f_{\theta} \) and \( L \)}

Assume that the model function \( f_{\theta}: \mathbb{R}^d \times \mathcal{T} \to \mathbb{R}^k \) and the loss function \( L: \mathbb{R}^k \times \mathcal{T} \to \mathbb{R} \) are Lipschitz continuous with constants \( K_f \) and \( K_L \), respectively. That is, for all \( x, y \in \mathbb{R}^d \) and \( T \in \mathcal{T} \):

\[
\| f_{\theta}(x, T) - f_{\theta}(y, T) \| \leq K_f \| x - y \|,
\]

and for all \( a, b \in \mathbb{R}^k \):

\[
| L(a, T^{\text{target}}) - L(b, T^{\text{target}}) | \leq K_L \| a - b \|.
\]

\noindent\textbf{Step 2: Bounding the Change in Loss Due to Noise \( \eta \)}

Consider the adversarially perturbed image \( I_{\delta} = I + \delta \), where \( \delta \) is crafted to minimize the loss:

\[
\delta = \arg\min_{\| \delta \| \leq \epsilon} L\left( f_{\theta}(I + \delta, T), T^{\text{target}} \right).
\]

When Gaussian noise \( \eta \sim \mathcal{N}(0, \sigma^2 I) \) is added, the input becomes \( I_{\delta, \eta} = I + \delta + \eta \). The change in loss due to \( \eta \) is:

\[
\Delta L = L\left( f_{\theta}(I + \delta + \eta, T), T^{\text{target}} \right) - L\left( f_{\theta}(I + \delta, T), T^{\text{target}} \right).
\]

Using the Lipschitz continuity of \( L \):

\[
| \Delta L | \leq K_L \left\| f_{\theta}(I + \delta + \eta, T) - f_{\theta}(I + \delta, T) \right\|.
\]

\noindent\textbf{Step 3: Computing the Expected Increase in Loss}

Applying the Lipschitz continuity of \( f_{\theta} \):

\[
\left\| f_{\theta}(I + \delta + \eta, T) - f_{\theta}(I + \delta, T) \right\| \leq K_f \| \eta \|.
\]

Thus, the change in loss is bounded by:

\[
| \Delta L | \leq K_L K_f \| \eta \|.
\]

Since \( \eta \) is a Gaussian random vector with zero mean and covariance \( \sigma^2 I \), the expected value of \( \| \eta \| \) is:

\[
\mathbb{E}[\| \eta \|] = \sigma \sqrt{2} \frac{\Gamma\left( \frac{d+1}{2} \right)}{\Gamma\left( \frac{d}{2} \right)} \approx \sigma \sqrt{d - \frac{1}{2}} \quad \text{for large } d.
\]

Therefore, the expected increase in loss is approximately:

\[
\mathbb{E}[ | \Delta L | ] \leq K_L K_f \sigma \sqrt{d}.
\]

\noindent\textbf{Step 4: Lower Bounding the Expected Increase in Loss}

Since \( \delta \) minimizes \( L(f_{\theta}(I + \delta, T), T^{\text{target}}) \) at the point \( I + \delta \), any perturbation \( \eta \) added to \( I + \delta \) is likely to increase the loss. Under the conjecture that \( L \) is convex around \( I + \delta \), the expected increase in loss due to \( \eta \) can be lower bounded using the curvature (second derivative) of \( L \):

\begin{align*}
    \mathbb{E}_{\eta} \left[ L\left( f_{\theta}(I + \delta + \eta, T), T^{\text{target}} \right) \right] \geq
    \\ L\left( f_{\theta}(I + \delta, T), T^{\text{target}} \right) + \frac{\sigma^2}{2} \lambda_{\text{min}},
\end{align*}

where \( \lambda_{\text{min}} \) is the smallest eigenvalue of the Hessian matrix \( \nabla_{I+\delta}^2 L(f_{\theta}(I + \delta, T), T^{\text{target}}) \).

\noindent\textbf{Conclusion:}

Adding Gaussian noise increases the expected loss by at least \( \frac{\sigma^2}{2} \lambda_{\text{min}} \), reducing the effectiveness of the adversarial perturbation. This result supports the conjecture that Gaussian noise disrupts the optimization achieved by the adversary, weakening the impact of adversarial attacks.

\subsection*{Second-Order Taylor Expansion of \( L \) Around \( I \)}

\subsubsection*{Discussion:}

\textbf{Step 1: Second-Order Taylor Expansion}

We expand the loss function \( L(f_{\theta}(I + \eta, T), T) \) around the point \( I \) using the second-order Taylor expansion:
\begin{align*}
    L(f_{\theta}(I + \eta, T), T) &= L(f_{\theta}(I, T), T) \\&+ \nabla_I L(f_{\theta}(I, T), T)^\top \eta \\&+ \frac{1}{2} \eta^\top \nabla_I^2 L(f_{\theta}(I, T), T) \eta + R_3
\end{align*}

where:
\begin{itemize}
    \item \( \nabla_I L(f_{\theta}(I, T), T) \) is the gradient of the loss with respect to the input \( I \).
    \item \( \nabla_I^2 L(f_{\theta}(I, T), T) \) is the Hessian matrix of second derivatives with respect to \( I \).
    \item \( R_3 \) is the remainder term of higher order \( O(\|\eta\|^3) \).
\end{itemize}

\noindent\textbf{Step 2: Expected Value of the Linear Term}

Since \( \eta \) is sampled from a zero-mean Gaussian distribution \( \eta \sim \mathcal{N}(0, \sigma^2 I) \), the expected value of the linear term becomes:
\[
\mathbb{E}_{\eta} \left[ \nabla_I L(f_{\theta}(I, T), T)^\top \eta \right] = \nabla_I L(f_{\theta}(I, T), T)^\top \mathbb{E}_{\eta}[\eta] = 0
\]

\noindent\textbf{Step 3: Expected Value of the Quadratic Term}

Next, we compute the expectation of the quadratic term:
\[
\mathbb{E}_{\eta} \left[ \eta^\top \nabla_I^2 L(f_{\theta}(I, T), T) \eta \right]
\]
Using the properties of Gaussian distributions, we know that for a symmetric matrix \( A \):
\[
\mathbb{E}_{\eta} \left[ \eta^\top A \eta \right] = \sigma^2 \operatorname{Tr}(A)
\]
Thus, the expected value of the quadratic term becomes:
\[
\mathbb{E}_{\eta} \left[ \eta^\top \nabla_I^2 L(f_{\theta}(I, T), T) \eta \right] = \sigma^2 \operatorname{Tr} \left( \nabla_I^2 L(f_{\theta}(I, T), T) \right)
\]

\noindent\textbf{Step 4: Neglecting the Remainder Term}

For small values of \( \sigma \), the remainder term \( R_3 \) is of order \( O(\sigma^3) \) and can be safely ignored. Thus, the approximation becomes:
\begin{align*}
    \mathbb{E}_{\eta} \left[ L(f_{\theta}(I + \eta, T), T) \right] &\approx L(f_{\theta}(I, T), T) \\&+ \frac{\sigma^2}{2} \operatorname{Tr} \left( \nabla_I^2 L(f_{\theta}(I, T), T) \right)
\end{align*}

\noindent\textbf{Step 5: Interpretation of the Trace Term}

The term \( \operatorname{Tr} \left( \nabla_I^2 L(f_{\theta}(I, T), T) \right) \) denotes the sum of the eigenvalues of the Hessian matrix, representing the overall curvature of the loss function with respect to the input. A larger trace value indicates higher curvature, suggesting greater sensitivity of the model to input perturbations. Reducing this sensitivity is crucial for enhancing the model's robustness.

\noindent\textbf{Step 6: Gaussian Noise as Regularization}

The additional term \( \frac{\sigma^2}{2} \operatorname{Tr} \left( \nabla_I^2 L(f_{\theta}(I, T), T) \right) \) functions as a regularizer, penalizing high curvature in the loss landscape. This encourages the model to learn smoother input-output mappings, thereby reducing its vulnerability to small perturbations, including adversarial attacks.

\noindent\textbf{Step 7: Connection to Tikhonov Regularization}

This regularization effect is conceptually similar to Tikhonov regularization, where a penalty proportional to the norm of the model parameters is added to the loss function. In our case, the penalty arises naturally from the Gaussian noise, encouraging robustness by flattening the loss landscape:
\begin{align*}
    \mathbb{E}_{\eta} \left[ L(f_{\theta}(I + \eta, T), T) \right] &\approx L(f_{\theta}(I, T), T) \\&+ \frac{\sigma^2}{2} \operatorname{Tr} \left( \nabla_I^2 L(f_{\theta}(I, T), T) \right)
\end{align*}

This smoothing effect reduces the model’s sensitivity to input perturbations, enhancing its robustness without compromising performance on clean data.

\newpage